%% file: main_arxiv.tex
\documentclass[runningheads]{llncs}

\usepackage[mobile]{eccv}

\usepackage{eccvabbrv}

\usepackage{graphicx}
\usepackage{booktabs}
\usepackage{xparse}
\usepackage{makecell}
\usepackage{arydshln}
\usepackage{wrapfig}
\usepackage{graphicx}
\usepackage{caption}
\usepackage[table]{xcolor}
\usepackage{siunitx}
\usepackage{multirow}
\usepackage{makecell}
\usepackage{enumitem}

\usepackage[accsupp]{axessibility}  % Improves PDF readability for those with disabilities.

\definecolor{lightgray}{gray}{0.92}

\usepackage[breaklinks,colorlinks,citecolor=eccvblue]{hyperref}

\usepackage{orcidlink}

\input{definitions}

\begin{document}

% ---------------------------------------------------------------
\title{
Foveated Diffusion:\\
\large Efficient Spatially Adaptive Image and Video Generation
}

\titlerunning{Foveated Diffusion}

\author{%
  Brian Chao$^{*}$ \and
  Lior Yariv$^{*}$ \and
  Howard Xiao\inst{1} \and
  Gordon Wetzstein\inst{1}%
}

\authorrunning{Chao, Yariv et al.}

\institute{Stanford University, USA}

\maketitle

{\renewcommand{\thefootnote}{}\footnotetext{$^*$ Denotes equal contribution.}}

\begin{abstract}
% v1
Diffusion and flow matching models have unlocked unprecedented capabilities for creative content creation, such as interactive image and streaming video generation. The growing demand for higher resolutions, frame rates, and context lengths, however, makes efficient generation increasingly challenging, as computational complexity grows quadratically with the number of generated tokens. Our work seeks to optimize the efficiency of the generation process in settings where the user's gaze location is known or can be estimated, for example, by using eye tracking. In these settings, we leverage the eccentricity-dependent acuity of human vision: while a user perceives very high-resolution visual information in a small region around their gaze location (the foveal region), the ability to resolve detail quickly degrades in the periphery of the visual field. Our approach starts with a mask modeling the foveated resolution to allocate tokens non-uniformly, assigning higher token density to foveal regions and lower density to peripheral regions. An image or video is generated in a mixed-resolution token setting, yielding results perceptually indistinguishable from full-resolution generation, while drastically reducing the token count and generation time. To this end, we develop a principled mechanism for constructing mixed-resolution tokens directly from high-resolution data, allowing a foveated diffusion model to be post-trained from an existing base model while maintaining content consistency across resolutions.
We validate our approach through extensive analysis and a carefully designed user study, demonstrating the efficacy of foveation as a practical and scalable axis for efficient generation.
Project website at \url{https://bchao1.github.io/foveated-diffusion/}.

\end{abstract}

\section{Introduction}
\label{sec:introduction}
\input{sections/introduction}

\section{Related Work}
\label{sec:related_work}
\input{sections/related_work}

\section{Method}
\label{sec:method}
\input{sections/method}

\section{Experiments}
\label{sec:results}

\input{sections/results}

\section{Conclusion}
\label{sec:conclusions}
\vspace{-5pt}
\input{sections/conclusions}

\section*{Acknowledgments}
We thank Ryan Po, Hansheng Chen, and Tong Wu for fruitful discussions.
Brian Chao and Howard Xiao are supported by Stanford Graduate Fellowships (SGF). Brian Chao is also supported by the NSF Graduate Research Fellowship Program (GRFP).
Compute resources were provided by the Marlowe cluster at Stanford University~\cite{marlowe2025}.

% ---- Bibliography ----
\bibliographystyle{splncs04}
\bibliography{main}

% ================================================================
% SUPPLEMENTARY MATERIAL
% ================================================================
\clearpage
\setcounter{section}{0}
\renewcommand{\thesection}{\Alph{section}}

\begin{center}
  {\LARGE\bfseries Supplementary Material}\\[0.5em]
  {\large Foveated Diffusion: Efficient Spatially Adaptive Image and Video Generation}
\end{center}

\vspace{1em}

\section*{Table of Contents}

\begin{enumerate}[itemsep=5pt]
    \item \textbf{Additional Implementation Details} \dotfill \pageref{supp_sec:implementation}
    \begin{enumerate}[label=\theenumi.\arabic*, start=1, itemsep=4pt, topsep=4pt]
        \item Image Generation \dotfill \pageref{supp_subsec:image_impl}
        \item Video Generation \dotfill \pageref{supp_subsec:video_impl}
    \end{enumerate}
    \item \textbf{Additional User Study Details} \dotfill \pageref{supp_sec:user_study}
    \begin{enumerate}[label=\theenumi.\arabic*, start=1, itemsep=4pt, topsep=4pt]
        \item Study Design and Participants \dotfill \pageref{supp_subsec:user_study_design}
        \item Study Setup and Images \dotfill \pageref{supp_subsec:user_study_setup}
        \item Procedure \dotfill \pageref{supp_subsec:user_study_procedure}
        \item Statistical Analysis \dotfill \pageref{supp_subsec:user_study_stats}
    \end{enumerate}
    \item \textbf{Additional Image Qualitative Results} \dotfill \pageref{supp_sec:image_qualitative}
    \begin{enumerate}[label=\theenumi.\arabic*, start=1, itemsep=4pt, topsep=4pt]
        \item Extended Image Generation Baseline Comparisons \dotfill \pageref{supp_subsec:image_baseline}
        \item Image Generation with Different Foveation Patterns \dotfill \pageref{supp_subsec:image_fov}
        \item Towards Saliency-guided Image Generation \dotfill \pageref{supp_subsec:image_saliency}
        \item Towards Bounding-box-guided Image Generation \dotfill \pageref{supp_subsec:image_bbox}
    \end{enumerate}
    \item \textbf{Additional Video Qualitative Results} \dotfill \pageref{supp_sec:video_qualitative}
    \begin{enumerate}[label=\theenumi.\arabic*, start=1, itemsep=4pt, topsep=4pt]
        \item Extended Video Generation Baseline Comparisons \dotfill \pageref{supp_subsec:video_baseline}
        \item Video Generation with Different Foveation Patterns \dotfill \pageref{supp_subsec:video_fov}
        \item Towards Saliency-guided Video Generation \dotfill \pageref{supp_subsec:video_saliency}
    \end{enumerate}
    \item \textbf{Discussion} \dotfill \pageref{supp_sec:discussion}
\end{enumerate}

\clearpage
\section{Additional Implementation Details}
\label{supp_sec:implementation}
\input{supp_sections/implementation}

\clearpage
\section{Additional User Study Details}
\label{supp_sec:user_study}
\input{supp_sections/user_study}

\clearpage
\section{Additional Image Qualitative Results}
\label{supp_sec:image_qualitative}
\input{supp_sections/image_qualitative}

\clearpage
\section{Additional Video Qualitative Results}
\label{supp_sec:video_qualitative}
\input{supp_sections/video_qualitative}

\clearpage
\section{Discussion}
\label{supp_sec:discussion}
\input{supp_sections/discussion}

\end{document}

%% file: definitions.tex
\newcommand{\latentspace}{\mathbb{R}^{c \times (h \cdot w)}}

\newcommand{\latentspacevideo}{\mathbb{R}^{c \times (f \cdot h \cdot w)}}
\newcommand{\gaussiannoise}{\mathcal{N}(0, I)}
\newcommand{\latentspacefoveated}{\mathbb{R}^{c \times L}}
\newcommand{\latentspaceLR}{\mathbb{R}^{c \times (\frac{h}{2} \cdot \frac{w}{2})}}

\newcommand{\spaceimage}{\mathbb{R}^{H \times W}}

\newcommand{\spacemasklatent}{\{0,1\}^{h \times w}}
\newcommand{\spacemaskimage}{\{0,1\}^{H \times W}}

\newcommand{\imageHR}{x^{\text{high}}}
\newcommand{\imageLR}{x^{\text{low}}}
\newcommand{\imageFOV}{x^{\text{fov}}}

\NewDocumentCommand{\latentHR}{O{t}}{z_{#1}^{\text{high}}}
\NewDocumentCommand{\latentLR}{O{t}}{z_{#1}^{\text{low}}}
\NewDocumentCommand{\latentFOV}{O{t}}{z_{#1}^{\mathrm{fov}}}
\newcommand{\latentclean}{z_0}
\newcommand{\latentnoise}{z_1}
\NewDocumentCommand{\latent}{O{t}}{z_{#1}}

\newcommand{\foveationmask}{M}
\newcommand{\encoder}{E}
\newcommand{\decoder}{D}

\newcommand{\ly}[1]{\textcolor{violet}{[LY: #1]}}

%% file: sections/introduction.tex
\begin{figure}[!t]
    \centering
    \includegraphics[width=\textwidth]{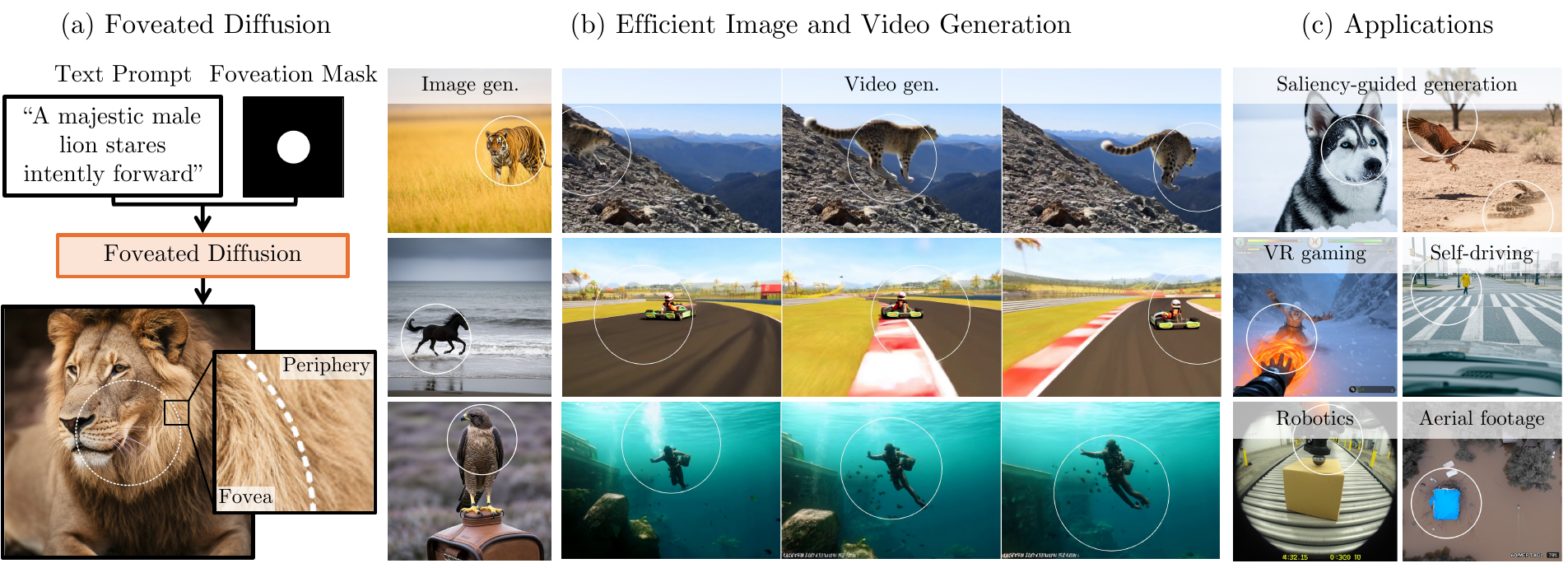}
    % \vspace{-17pt}
    \caption{\textbf{Foveated Diffusion.} (a) Given user-specified masks and text prompts as input, our method generates foveated content using fewer tokens than full high-resolution generation, resulting in faster inference while maintaining comparable perceptual quality.
    (b, c) Foveated Diffusion is well suited for tasks where salient regions require high-resolution synthesis, while peripheral regions can be generated at a lower resolution.}
    % \vspace{-3pt}
    \label{fig:teaser}
\end{figure}

Interactive image and streaming video generation place strict demands on the frame rates of emerging diffusion and flow matching models used for this purpose~\cite{bruce2024genie,alonso2024diffusion,che2024gamegen,decart2024oasis,feng2024matrix,jin2024pyramidal,kodaira2025genie3,song2025history,valevski2024diffusion,weng2024art,yu2025gamefactory,henschel2025streamingt2v,wu2025spmem,po2025long,zhang2025frame,yin2025slow}. At the same time, demands on image resolutions and video frame or context lengths are also growing. How can we generate an ever-increasing number of tokens at fast frame rates when the computational complexity of the attention mechanism in modern diffusion transformers (DiTs) \cite{peebles2023dit} grows quadratically with the token sequence length?

Our work builds on an intuitive insight that answers this question: ultimately, a human observes the generated content, so why not exploit the unique characteristics of the human visual system to generate the content in a computationally efficient, perceptually motivated manner? Specifically, we build on the concept of foveation --- humans are able to perceive very high-resolution visual information in a small region around their gaze location (the foveal region) but their ability to resolve detail rapidly degrades in the visual periphery~\cite{anstis1974chart,weymouth1958visual}. 

With this work, we introduce the concept of \emph{Foveated Diffusion} and develop a practical framework for post-training existing image or video generation models for foveated visual generation. Our framework starts with a foveation mask that guides the spatial layout of non-uniformly distributed tokens over the image or video frame that we wish to generate. Our key idea is eccentricity-dependent token allocation: given a foveation mask that defines the high-acuity foveal region, we allocate higher token density near the fovea and progressively fewer tokens toward the periphery, enabling spatially adaptive computation aligned with human perceptual sensitivity.
Using a foveated token layout, we follow standard diffusion or flow-matching procedures to generate an image from Gaussian noise and conditioning text prompts (see Fig.~\ref{fig:teaser}); the key difference between Foveated Diffusion and conventional methods is that we operate with a significantly reduced set of tokens during denoising at all times, achieving substantial computational savings. We develop a simple yet highly effective mixed-resolution tokenization scheme, accompanied by a suitable modification of Rotary Positional Embeddings (RoPE)~\cite{su2024roformer,wu2025crpa}, along with a post-training strategy that transforms high-resolution pretrained models into foveated generative models. Together, these contributions establish a principled framework that preserves cross-resolution content consistency while achieving significant speedup.

Our approach is inspired by foveated rendering~\cite{geisler1998foveated,guenter2012foveated3d,patney2016gazeVR}, a standard technique widely used in traditional computer graphics. Foveated Diffusion and rendering share the idea of leveraging the user's gaze location and a model of eccentricity-dependent acuity to reduce computation in the visual generation process. The key difference is that foveated rendering accelerates modules of the traditional graphics pipeline, such as the geometry and shading engines, whereas our approach seeks to achieve similar benefits for modern DiT-based diffusion and flow-matching models. While similar in spirit, these two approaches to foveated content creation differ substantially in their methods. 

In summary, we propose a perceptually motivated framework for computationally efficient and spatially adaptive image and video generation. Our approach is backed by an extensive set of evaluations, including a detailed analysis of the compute--quality trade-off in various settings as well as a user study. \\
Our key contributions are as follows:
\begin{itemize}
    \item We introduce of the concept of \emph{Foveated Diffusion}: a perceptually motivated, mixed-resolution diffusion algorithm for efficient image and video generation.
    \item We present a principled approach for tokenization, training, and inference of DiT-based generative models using spatially adaptive mixed-resolution tokens, providing cross-resolution content consistency by design.
    \item We demonstrate significant speedups of up to $2\times$ and $4\times$ for image and video generation, respectively, while preserving perceptual quality, validated through a carefully designed user study and visual quality metrics.
\end{itemize}

%% file: sections/related_work.tex
% show foveated rendering idea, visual eccentricity
%\bc{put way less emphasis on VR} \\
%\hx{Not exactly sure where to put these, but here are foveated vision/perception references. Heuristic-based or learned :~\cite{itti2002model, ba2014multiple}; policy-based:~\cite{mnih2014recurrent, wang2025emulating}}

\paragraph{\bf{Foveation for Computer Vision and Rendering.}} Decades of vision research have shown that visual acuity decreases rapidly with retinal eccentricity, i.e. distance from the fovea, where spatial resolution is highest \cite{curcio1990topography, rovamo1979magnification, watson2014formula, geisler1998foveated}. As a result, the human visual system processes central vision at significantly higher spatial precision than the periphery. 
% This inherent spatial asymmetry motivates allocating computation non-uniformly across the image, concentrating capacity near the gaze location while reducing resolution in peripheral regions.

Real-time rendering systems exploit this eccentricity-dependent resolution of human vision by allocating higher spatial resolution near the gaze location and lower resolution in peripheral regions. When combined with real-time eye tracking, such foveated rendering systems reduce bandwidth and compute substantially, enabling interactive rendering at a fraction of the full-resolution cost while maintaining comparable perceptual quality \cite{guenter2012foveated3d, patney2016gazeVR, levoy1990gazevolume, reddy2002perceptually, stengel2016adaptive, sun2017perceptuallyLF, kaplanyan2019deepfovea, weier2016foveated, tariq2022noise}. More recently, foveation has been applied to neural rendering and novel view synthesis~\cite{shi2024sceneFoVNeRF, franke2025vrsplat, deng2022fovnerf} to accelerate the rendering of Neural Radiance Fields (NeRFs) \cite{mildenhall2020nerf} and Gaussian splats \cite{kerbl3Dgaussians} for immersive displays. In computer vision and robotics, foveation has also been used to improve efficiency in neural network architectures or perception tasks \cite{minut2000face, bandera1989foveal, killick2023foveation}, such as mixed-resolution tokenization of vision transformers \cite{jonnalagadda2021foveater, ronen2023mixedrestoken, schmidt2025segment, havtorn2023msvit} and robot policy learning \cite{kerrj2025eyerobot, chuang2025lookfocusactefficient}.

However, while foveation has been extensively explored across rendering and perception pipelines, 
% modern generative models offer no mechanism for spatially adaptive computation. As a result, 
it has not yet been realized in generative modeling, despite their increasing capabilities in immersive and interactive visual generation. This gap motivates the need for a generative framework that can allocate capacity according to visual eccentricity.

\paragraph{\bf{Efficient Visual Generation.}} Diffusion models have fundamentally reshaped visual generative modeling, setting new standards in photorealism, diversity, and controllability for both images and videos. While early diffusion models leverage U-Net backbones \cite{rombach2022latent}, Diffusion Transformer (DiT)-based architectures have emerged as the dominant paradigm for scalable, high-fidelity generation \cite{blackforestlabs2025flux2klein, wan2025wan, kong2024hunyuanvideo, esser2024flow, peebles2023dit}. However, the computational cost of DiTs is quadratic with respect to the input token count due to the expensive self- and cross-attention mechanisms \cite{vaswani2017attention}. This fundamental limitation of transformer architectures severely constrains context length, leading either to degraded visual consistency under fixed compute and memory budgets or to prohibitive computational costs for immersive, high-fidelity, long-form generation. 

There have been significant efforts in improving the computational efficiency of the attention mechanism, including various attention variants that reduce the algorithmic complexity \cite{li2025radial, xia2025trainingfree, zhan2025bidirectional, katharopoulos202linear, wang2020linformer, choromanski2021performer, beltagy2020longformer}, hardware-aware optimization \cite{zhang2025spargeattn, zhang2025vsa, zhang2025STA, xi2025sparsevideogen, dao2022flashattention}, KV-caching \cite{kwon2023paged, shazeer2019mqa}, etc. Another orthogonal axis of research aims to simply reduce the effective token count while maintaining high image quality. Token merging methods \cite{bolya2022tome, lee2024video, chen2025comeconfidenceguidedtokenmerging} identify redundant tokens at each layer of a Diffusion Transformer (DiT) and merge similar tokens according to a predefined heuristic or importance metric. While originally developed for vision transformers in recognition and perception tasks, they have recently been shown to effectively reduce token counts for generative models as well \cite{lu2025toma, haurum2024agglomerative, bolya2023token, kim2024tokenfusion, wu2025importance, lee2025local, fang2025attend}. Recent training-free mixed-resolution denoising methods \cite{jeong2025upsample, wu2025crpa, tian2025bottleneck} downsample or upsample tokens during the diffusion process using fixed importance metrics such as entropy or saliency to reduce token counts for efficient generation. However, directly applying standard denoising to mixed-resolution tokens requires carefully tuned noise schedules and re-noising strategies to preserve diffusion noise statistics and maintain global content structure. These procedures are brittle; without them, mixed-resolution generation leads to structural inconsistencies and cross-scale artifacts, as we demonstrate in Sec. \ref{sec:results}. In addition, existing approaches rely on multi-stage pipelines in which a low-resolution image first establishes global layout, followed by progressive token upsampling. Such designs complicate the diffusion trajectory and hinder compatibility with real-time generation and model distillation. 

Although all these methods significantly improve efficiency in visual generation, they ignore a key characteristic of visual perception: human visual acuity decreases sharply with eccentricity. These methods focus on reconstructing high-resolution imagery everywhere and treat all spatial regions uniformly. However, because generated images are intended for human observers, generation should be optimized for perceptual relevance rather than uniform pixel fidelity. In contrast, our Foveated Diffusion pipeline leverages this principle by embedding spatially adaptive token allocation directly into the diffusion process given a predetermined foveation mask. By concentrating computation in high-acuity regions and sparsifying peripheral regions, we depart from uniform-resolution synthesis and achieve perceptually aligned, computationally efficient generation.

%% file: sections/method.tex
In this section, we first review the basic concepts of foveated rendering in traditional graphics, as well as standard diffusion and flow-matching models in Sec. \ref{subsec:method_preliminaries}. We then introduce our Foveated Diffusion framework (Fig. \ref{fig:pipeline}) and explain its tokenization, inference, and training pipelines in Sec. \ref{subsec:method_foveated_diffusion}.

\subsection{Preliminaries}
\label{subsec:method_preliminaries}

\subsubsection{Foveated Rendering.}\label{subsubsec:fovrend}
Foveated rendering refers to the spatially adaptive computation where computational resources are allocated unevenly across the image according to a specified user gaze location. Modern real-time graphics systems leverage eye-tracking to render high-resolution imagery in the foveal regions while aggressively reducing shading, rasterization, or sampling rates in the peripheral regions \cite{bandera1989foveal, geisler1998foveated, guenter2012foveated3d, weier2016foveated, patney2016gazeVR, kaplanyan2019deepfovea}.

Formally, we define a binary foveation mask $\foveationmask \in \spacemaskimage$ constructed from visual eccentricity, where $M(i,j)=1$ denotes high-resolution (HR) regions near the 
fovea and $M(i,j)=0$ denotes low-resolution (LR) peripheral regions. The rendering quality is concentrated in the HR region near the fixation point (center of foveation), and progressively reduced toward the periphery. Most importantly, in foveated rendering, the scene content is unknown a priori and the foveation mask is known via gaze.

We denote $x^{\text{high}} \in \mathbb{R}^{3 \times H \times W}$ as the underlying high-resolution content, and 
$x^{\text{low}} \in \mathbb{R}^{3 \times (H/d) \times (W/d)}$ as the underlying low-resolution content, 
where $d$ is the spatial downsampling factor. In this paper, we define $d=2$, allowing $4\times$ computational gain in the periphery. During foveated rendering, only pixels $(i,j)$ with $M(i,j)=1$ are synthesized at high resolution from $x^{\text{high}}$ (the foveal region), while pixels with $M(i,j)=0$ are synthesized from $x^{\text{low}}$ (the peripheral region). Thus, computation is performed exclusively on the masked regions at their respective resolutions, rather than producing full high- and low-resolution renderings. 
Composing the final foveated image in pixel space is simply achieved by blending, that is:
\begin{align}\label{eq:fov_image_merge}
    x_{\text{fov}} 
    = M \odot x^{\text{high}}
    + (1 - M) \odot \mathrm{Up}(x^{\text{low}}),
\end{align}
where $\mathrm{Up}(\cdot)$ denotes the spatial upsampling operator, and $\odot$ denotes elementwise multiplication.

\subsubsection{Diffusion Models.} Diffusion models \cite{ho2020ddpm, ho2020denoising, song2020score} define a generative process that gradually transforms samples from an easy-to-sample distribution (i.e. Gaussian) into data samples via a learned reverse-time process. Modern large-scale diffusion models operate in a compressed latent space to improve computational efficiency \cite{peebles2023dit, rombach2022latent}. Given an image or a video, a variational autoencoder (VAE) \cite{Diederik_2019}, consisting of an encoder $\encoder$ and a decoder $\decoder$, maps it into a latent representation \( \latentclean \in \latentspace \) (images) or \( \latentclean \in \latentspacevideo \) (videos, with additional frame dimension $f$), where the diffusion process is defined.

\iffalse
In latent diffusion models, a noisy latent \( \latent{t} \) at timestep \( t \in [0,1] \) is obtained by interpolating between the clean latent and Gaussian noise:
\begin{align}
    \latent{t} = (1 - t)\latentclean + t \latentnoise, 
    \qquad \latentnoise \sim \mathcal{N}(0, I).
\end{align}
A neural network parameterized by \( \theta \) is trained to predict the denoising direction, enabling iterative sampling from noise to data.
\fi

\subsubsection{Flow Matching.} Flow matching \cite{lipman2023flow} reformulates diffusion as a continuous-time optimal transport problem between the data distribution and a simple prior, usually a Gaussian distribution $\gaussiannoise$. Instead of learning to predict noise or score functions, flow matching learns a velocity field that deterministically transports samples along straight-line paths in latent space.

Specifically, given a data sample $\latentclean$ and a noise sample $\latentnoise \sim \gaussiannoise$, the noise-to-data path is defined via a linear interpolation:
\begin{align}
    \latent[t]= (1 - t)\latentclean + t \latentnoise.
    \label{eq:noisy_latent}
\end{align}
A neural network \( v_\theta(\latent[t], t) \) is optimized to predict its corresponding velocity field: \vspace{-10pt}
\begin{align}
    \frac{d}{dt} z_t = \latentnoise - \latentclean,
    % \dot{z}_t = \latentnoise - \latentclean.
\end{align}
Therefore, the training objective is to minimize
\begin{align}
    \mathbb{E}_{\latentclean, \latentnoise, t}
    \left[
        \left\|
            v_\theta(\latent[t], t) - (\latentnoise - \latentclean)
        \right\|_2^2
    \right].
    \label{eq:pred_flow}
\end{align}

At inference time, data samples are generated through sampling $\latent[1]$ and solving the flow ODE: $\frac{d}{dt} z_t = v_\theta(\latent[t], t)$. Flow matching yields faster convergence and more stable training compared to score-based diffusion models.

Almost all modern diffusion and flow matching models are built on top of the Diffusion Transformer (DiT) architecture \cite{peebles2023dit}. The computational efficiency of such generative models is therefore quadratically related to the number of tokens processed due to the expensive attention \cite{vaswani2017attention} and MLP operations in the DiT. This motivates our method, which generates images and videos using a reduced set of tokens where the low-resolution tokens are specified by spatial or spatiotemporal foveation masks, while preserving perceptual image quality.

% preliminaries, discuss token efficiency, and how foveated diffusion reduces token counts, diffusion->flow matchign

\subsection{Foveated Diffusion}
\label{subsec:method_foveated_diffusion}

\begin{figure}[t!]
    \centering
    \includegraphics[width=\textwidth]{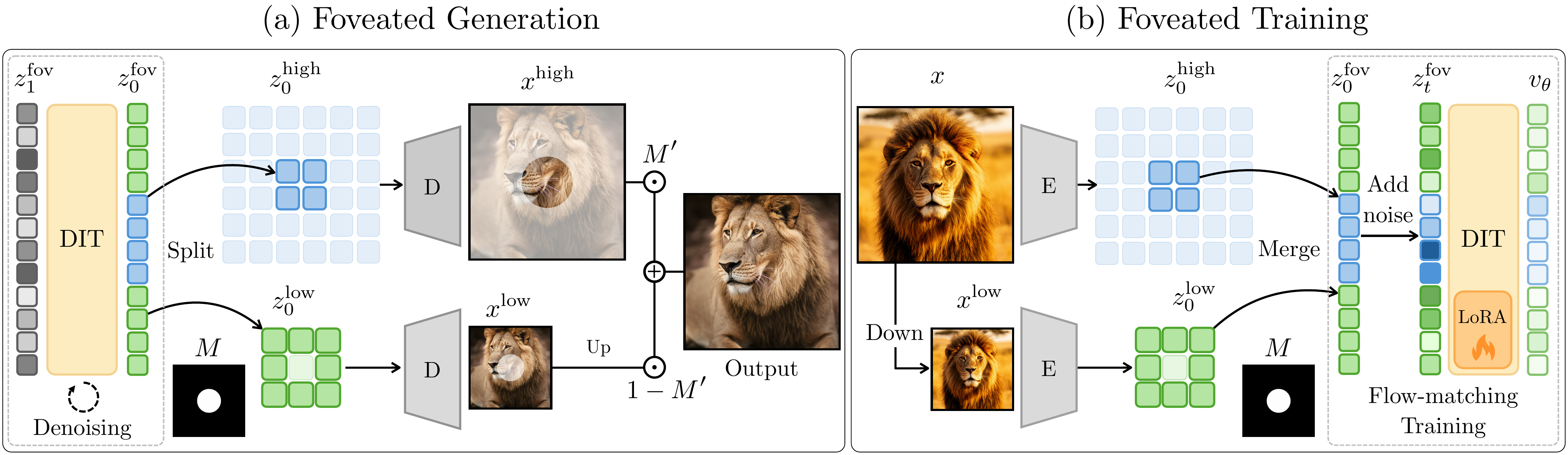}
    \vspace{-15pt}
    \caption{\textbf{The Foveated Diffusion Pipeline.} In Foveated Generation (a), we iteratively denoise a foveated token sequence of reduced length instead of the full high-resolution sequence. The resulting tokens $\latentFOV[0]$ are split into high- and low-resolution grids, decoded by the VAE, and blended using a user-specified foveation mask. We employ Foveated Training (b) to adapt pretrained DiTs to foveated token sequences using low-rank adaptation (LoRA) \cite{hu2022lora}. The image and its downsampled version are independently encoded by the VAE encoder and merged into a clean foveated token sequence for flow-matching training.}
    \label{fig:pipeline}
\end{figure}

\vspace{-3pt}
To achieve true computational savings in foveated visual generation, we introduce \emph{Foveated Diffusion}, a principled training and generation framework that enables diffusion or flow-matching models to directly generate spatially foveated images and videos with reduced token complexity. We describe our pipeline using latent-space image generation models here, but this concept applies equally to video generation models, as can be seen in Sec. \ref{sec:results}.

\vspace{-3pt}
\subsubsection{Foveated Tokenization. }\label{subsubsc:fov_token}
Let the latent space of a high-resolution image be $\latentspace$. The VAE encodes and patchifies each image $x\in \mathbb{R}^{3 \times H \times W}$ into a sequence of $h \times w$ tokens with feature dimension $c$. Standard DiTs perform training and generation directly on this full set of $h \cdot w$ tokens. 
In Foveated Diffusion, we are given a foveation mask $\foveationmask \in \spacemasklatent$ that specifies the spatial locations where high-resolution tokens are retained; meanwhile, peripheral regions are represented with fewer tokens to reduce the sequence length and computational complexity. Consequently, we operate entirely in the \emph{foveated token space} $\latentspacefoveated$ where the token sequence has a variable length $L \ll h \cdot w$. In our setting, a single low-resolution token represents the spatial area of a $2\times 2$ block of high-resolution tokens. This results in a total sequence length of $L = m + (h \cdot w - m) / 4$, where $m$ is the number of effective tokens in the mask $M$.
% Fig.~\ref{fig:pipeline} illustrates our mixed-resolution tokenization scheme.
This approach directly parallels foveated rendering in traditional graphics (see Sec. \ref{subsubsec:fovrend}), where shading and rasterization are computed asymmetrically based on a user-specified mask to achieve computational savings.

% \subsubsection{Foveated Tokenization. }
% Suppose the latent space of a high-resolution image is $\latentspace$, where the VAE encodes and patchifies each image $x\in \mathbb{R}^{3 \times H \times W}$ into a sequence of $h \times w$ tokens with feature dimension $c$. Standard DiTs perform tokenization, training, and generation directly on this full set of $h \cdot w$ tokens. In Foveated Diffusion, we instead operate entirely in a \emph{foveated token space} $\latentspacefoveated$. The foveated token sequence has variable length $L \ll h \cdot w$, where $L$ is determined by a predefined foveation mask $\foveationmask \in \spacemasklatent$. The mask specifies the retained high-resolution token locations, while peripheral regions are represented with fewer tokens, thereby reducing sequence length and computational complexity.
% In our setting, one low-resolution token represents the spatial area of $2\times 2$ high-resolution tokens, resulting with a new sequence length of $L = m + (h \cdot w - m) / 4$, where $m$ is the number of active tokens in the mask $M$. We refer the reader to Fig. \ref{fig:pipeline} for illustration of our mixed-resolution tokenization.
% This setting directly parallels the case of foveated rendering (see Sec. \ref{subsubsec:fovrend}) in traditional graphics, where shading and rasterization is computed asymmetrically given a user-specified foveation mask to achieve computational savings.

\subsubsection{Foveated Generation. }
\label{subsubsec:fovinfer}
Foveated generation is performed by sampling Gaussian noise $\latentFOV[1] \sim \mathcal{N}(0, I)$ in the foveated token space $\latentspacefoveated$ at a reduced sequence length $L$. The foveated token sequence is then iteratively denoised from $t=1$ to $t=0$,
producing a clean foveated token sequence $\latentFOV[0] \in \latentspacefoveated$. This procedure can be done in a completely training-free setting using a pretrained generative model, which we refer to as \emph{Na\"ive Mixed-Resolution Denoising}.

% In contrast to other methods that target full high resolution generation with mixed-resolution tokens \cite{jeong2025upsample, wu2025crpa}, our procedure, which we refer to as \emph{Na\"ive Mixed-Resolution Denoising}, assigns a single global noise level per timestep to all foveated tokens and performs joint denoising in the foveated token space, which completely removes the need for re-noising or specially crafted noise schedules 

To obtain a full-resolution image, we first partition the clean foveated token sequence $\latentFOV[0]$ into 
high-resolution and low-resolution components:
\begin{align}
    (\latentHR[0], \latentLR[0]) 
    = \mathrm{Split}(\latentFOV[0], M).
    \label{eq:split_fov_latent}
\end{align}
We then decode each subset separately with the VAE decoder $\decoder$:
\begin{align}
    \imageHR = \decoder(\latentHR[0]), 
    \qquad
    \imageLR = \decoder(\latentLR[0]).
    \label{eq:vae_decode}
\end{align}

%\ly{why do we need to define this equation again? the only difference is that here we use $M'$, but we can simply replace the purpose of $M'$ to be in latent space and just refer to the equation in 1, and write: "... and blended with the high-resolution decoding to form the final image $\imageFOV$", following the procedure defined in Eq. \ref{eq:fov_image_merge}."}

The decoded low-resolution image is spatially upsampled to the original 
spatial resolution and blended with the high-resolution decoding to form the 
final image using the upsampled latent foveation mask $M'=\mathrm{Up}(\foveationmask)\in \spaceimage$:
\begin{align}
    \imageFOV
    = M' \odot \imageHR
    + (1 - M') \odot \mathrm{Up}(\imageLR),
    \label{eq:image_merge}
\end{align}
where $\mathrm{Up}(\cdot)$ denotes spatial upsampling and $\odot$ denotes 
elementwise multiplication. The full generation pipeline is illustrated in Fig. \ref{fig:pipeline}-(a).

\begin{wrapfigure}[12]{r}{0.3\textwidth} % r/l for right/left; width of the wrapped box
  \centering
  \vspace{-0.1\baselineskip} % users can freely adjust top position by changing this
  \includegraphics[width=\linewidth]{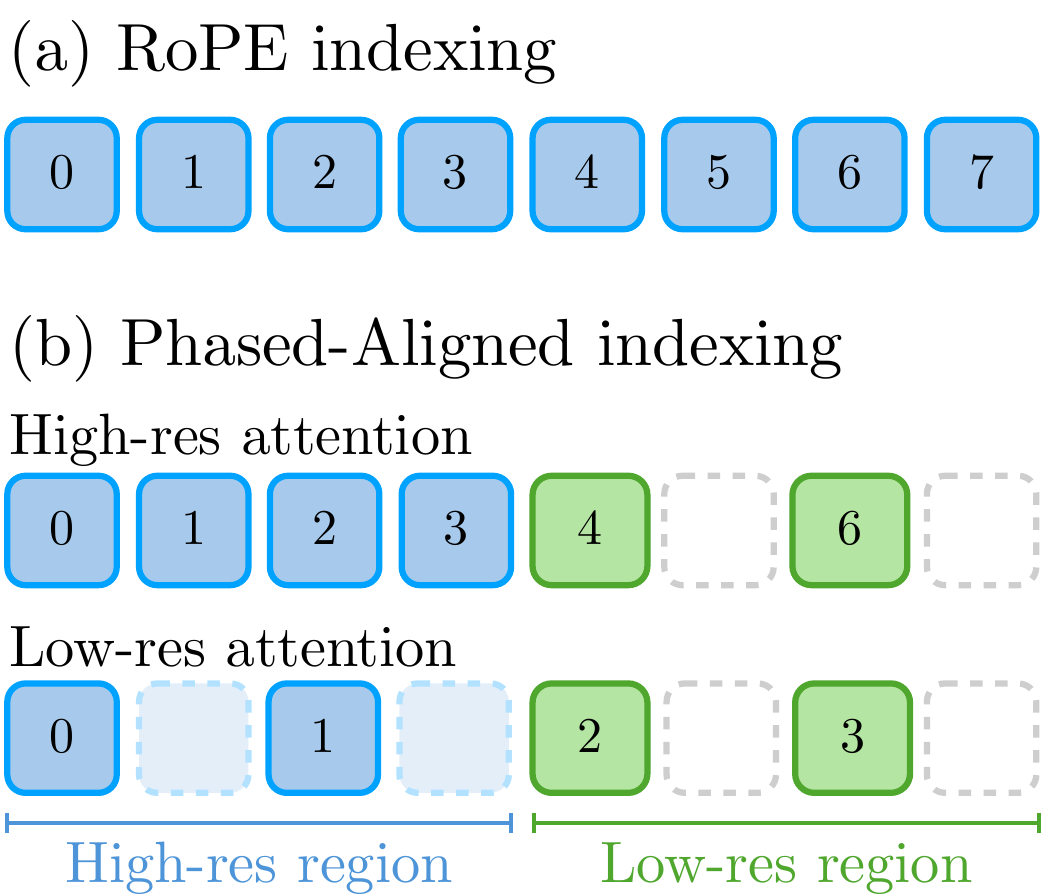}
  \caption{\textbf{Adapting RoPE for mixed-resolution attention \cite{wu2025crpa}.}}
  \label{fig:crpa}
  \vspace{-1.0\baselineskip} % (optional) tighten space below
\end{wrapfigure}
\paragraph{Mixed-Resolution RoPE.} 
Standard Rotary Positional Embedding (RoPE) \cite{su2024roformer} typically assumes a uniform grid with fixed-phase spacing (Fig. \ref{fig:crpa}(a)). However, because Foveated Diffusion introduces mixed-resolution tokenization, we must modify the RoPE indexing accordingly.
% Changing the tokenization input to the DiT requies modifications to the Rotary Positional Embedding \cite{su2024roformer} (RoPE). \edit{Vanilla RoPE indexing follows a uniform grid with fixed-phase spacing,} as illustrated in Fig. \ref{fig:crpa}(a).
% \edit{However, since Foveated Diffusion splits tokens into low- and high-resolution regions, RoPE indexing need to be adjusted to remove artifacts \cite{wu2025crpa}}. 
Therefore, we follow Wu et al.\cite{wu2025crpa} and align the key RoPE phases with query RoPE phases based on their corresponding token resolutions. Specifically, when computing attention with high-resolution query tokens, we sub-sample low-resolution key tokens from the full-resolution tokens. For attention with low-resolution query tokens, we subsample high-resolution key tokens and normalize their RoPE indices to the low-resolution grid. Please see illustration in Fig. \ref{fig:crpa}(b) or refer to Wu et al. \cite{wu2025crpa} for more details. 
% Modification to Rotary Positional Embeddings.
% Since the DiT was never trained on mixed-resolution tokens and the positional embeddings of tokens are assumed to lie on a uniform rectangular grid with fixed phase increments, na\"ively re-sampling RoPE \cite{su2024roformer} to adapt to mixed-resolution tokens leads to severe visual degradation as presented in \cite{wu2025crpa}. We follow the RoPE adaptation as suggested by Cross-Resolution Phase-Aligned Attention (CRPA) \cite{wu2025crpa} for both our method and the na\"ive mixed-resolution baseline. In short, CRPA works by mapping all query and key embedding positions onto the query's specific reference resolution grid during each attention call, which ensures that equal physical distances always produce identical phase increments in RoPE. Please refer to the CRPA paper \cite{wu2025crpa} for more details.

\paragraph{Failure of Na\"ive Mixed-Resolution Denoising.}
\label{sec:failure}

\begin{wrapfigure}{r}{0.3\textwidth} % r/l for right/left; width of the wrapped box
  \centering
  \vspace{-1.8\baselineskip} % users can freely adjust top position by changing this
  \includegraphics[width=\linewidth]{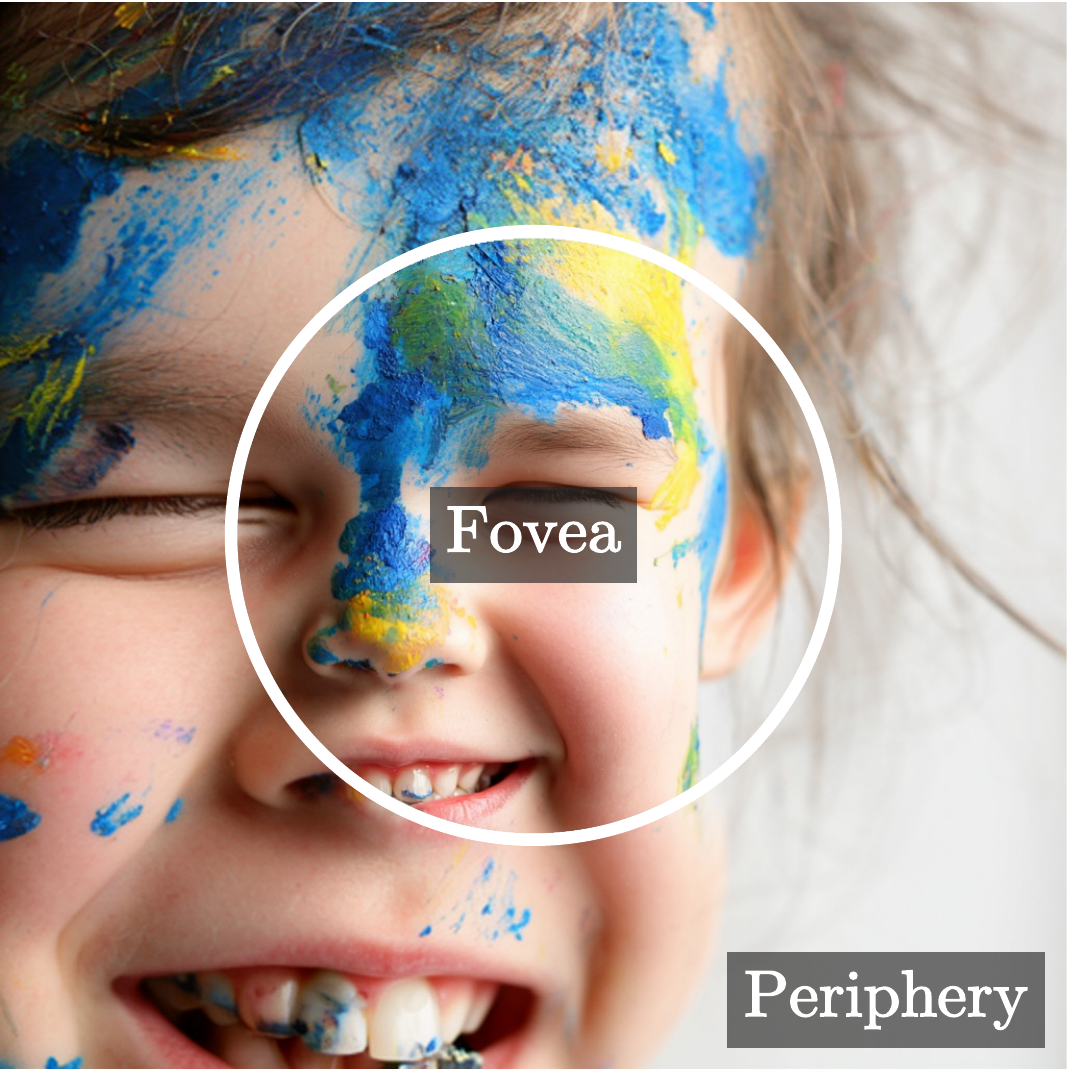}
  \caption{\textbf{Failure of Na\"ive mixed-resolution denoising.}}
  \label{fig:failure}
  \vspace{-2.0\baselineskip} % (optional) tighten space below
\end{wrapfigure}

\begin{figure}[!t]
    \centering
    \includegraphics[width=\textwidth]{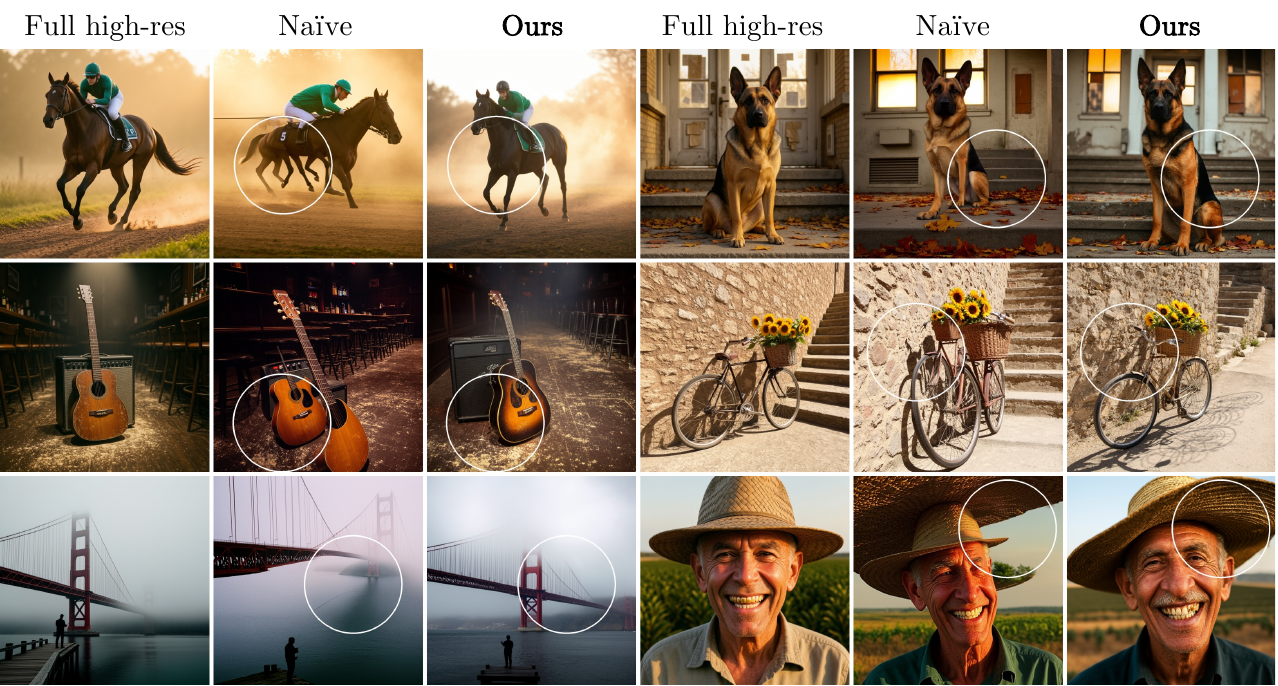}
    \vspace{-15pt}
    \caption{\textbf{Qualitative comparison for image generation.} Our method yields perceptually indistinguishable results from full high-resolution synthesis, whereas the naïve baseline introduces scale inconsistencies and structural artifacts across mixed-resolution regions. The high-resolution regions (fovea) are delineated with white borders.}
    \label{fig:baseline_image}
\end{figure}

A pretrained DiT is not, by default, compatible with a mixed-resolution, or foveated, token layout since tokens and positional embeddings are defined to be on a uniform grid. Even after adapting RoPE to handle mixed-resolution tokens, the low- and high-resolution regions still exhibit significant scale and structural inconsistencies, frequently resulting in duplicated objects or multiple entities fused together unnaturally (Fig. \ref{fig:failure}, and Fig. \ref{fig:baseline_image}). These findings suggest that high-quality foveated generation cannot be achieved with training-free mixed-resolution denoising, and a more principled approach is required to achieve our objective.

\iffalse
Consequently, if a model has never been adapted to mixed-resolution inputs, high-resolution (fovea) and low-resolution tokens (periphery) tend to be denoised as if they were independent denoising paths \ly{I don't think it is denoised as independent denoising paths cause the content is the same (you have the same child...) it is only the scale and boundaries are the big issues. let's just say:  "Consequently, if a model has never been adapted to mixed-resolution inputs, we can face with a scale imbalance between the high-resolution (fovea) and low-resolution (periphery) regions, and the boundary between the two is pretty noticeable." (remove also the following sentence, or combine/add what is required) }. When decoded and blended, the independently generated regions exhibit both semantic and scale inconsistencies, frequently resulting in duplicated objects or multiple entities fused together unnaturally, as shown in Fig. \ref{fig:failure}. These artifacts indicate that foveated generation requires explicit adaptation to enforce mixed-resolution during the denoising process \ly{perhaps let's repharse this more strongly to emphasize out contribution: "These artifacts indicate that foveated generation is not trivial to achieve via a mixed-resolution denoising, and a more principled approach is required for our objective.}.
\fi

\subsubsection{Foveated Training.}
\label{sec:training}
To address the aforementioned failure, we design an effective post-training procedure in the foveated token space by constructing foveated training targets that are mixed-resolution-consistent by design, as shown in Fig. \ref{fig:pipeline}-(b). Given a high-resolution training image $x$, we first form two latent token sequences using the VAE encoder $\encoder$. The high resolution token sequence is:
\begin{equation}
    \latentHR[0] = \encoder(x) \in \latentspace
    \label{eq:HR_latents}
\end{equation}
The low-resolution token sequence is obtained by bicubically downsampling the image and encoding it:
\begin{equation}
    \latentLR[0] = \encoder(\mathrm{Down}(x)) \in \latentspaceLR
    \label{eq:LR_latents}
\end{equation}

We then construct a clean foveated target token sequence by merging tokens from $\latentHR[0]$ and $\latentLR[0]$ according to the foveation mask:
\begin{equation}
\label{eq:fov_merge}
\latentFOV[0] = \mathrm{Merge}(\latentHR[0], \latentLR[0], M) \in \latentspacefoveated.
\end{equation}
By construction, both $\latentHR[0]$ and $\latentLR[0]$ are derived from the same underlying image content, and thus $\latentFOV[0]$ defines a single coherent target token sequence for mixed-resolution denoising. This clean foveated token sequence $\latentFOV[0]$ exactly corresponds to the foveated image $\imageFOV$ through Equations $\ref{eq:split_fov_latent}$ to $\ref{eq:image_merge}$.

We train our Foveated Diffusion model using the standard flow matching objective \cite{lipman2023flow}. Concretely, for a sampled timestep $t$ and noise $\latentnoise^{\text{fov}} \sim \gaussiannoise$, we generate a noisy foveated token sequence $\latentFOV[t]$ from $\latentFOV[0]$ following the flow-matching parameterization (Eq. \ref{eq:noisy_latent}) and optimize the model to predict the corresponding target velocity (Eq. \ref{eq:pred_flow}). Most importantly, all computations are performed on the variable-length $L$ foveated token sequence.

\subsubsection{Foveation Masking Strategies.} \label{subsub:foveation_mask_stragegy} Our Foveated Training procedure is agnostic to how the foveation masks $\foveationmask$ are designed, as it is simply a user-specified binary mask indicating the locations of high-resolution tokens. The form of the masks for training is therefore flexible and completely task-dependent. Specifically, we present two variants in the main paper: 
\textit{randomized masks}, which produce a generative model agnostic to the specified foveation, allowing the gaze center to be shifted to arbitrary image regions; and 
\textit{saliency-guided masks}, which encourage the foveal region to encompass salient objects in the scene, reflecting the regions where viewer attention is most likely to be directed. We additionally show \textit{bounding-box masks} results in the supplementary materials.

Notably, training with different masking strategies does not require any modification to the model architecture or to the training objective; it only involves changing the foveation masks.

%% file: sections/results.tex
\subsection{Implementation Details}

% datasets processing, base model, tokenAE architecture

For image and video generation, we adopt pretrained Diffusion Transformers (DiTs) as base models and fine-tune them using our Foveated Diffusion framework. For image generation, we fine-tune FLUX.2 Klein 4B \cite{blackforestlabs2025flux2klein} on the Aesthetic-Train-V2 dataset~\cite{zhang2025diffusion4k, zhang2025ultrahighresolutionimagesynthesis}. We randomly sample 90k images for training and reserve 10k prompt--image pairs for evaluation. For video generation, we fine-tune Wan2.1 1.3B \cite{wan2025wan} on Vchitect-T2V-Dataverse \cite{fan2025vchitect,si2025RepVideo}, excluding 200 prompts to serve as test samples. During training, we randomly sample a circular foveation mask for each image or a random foveation path for video to simulate diverse foveation patterns.
For a fair comparison, the full-resolution and naïve mixed-resolution baselines use the same base models and training data, but without Foveated Diffusion training. All models are fine-tuned using Low-Rank Adaptation (LoRA) \cite{hu2022lora} with rank 32. Experiments are conducted on NVIDIA H100 GPUs. Please refer to the supplementary materials for more details.

\subsection{Results}

\begin{wrapfigure}[11]{r}{0.3\textwidth} % r/l for right/left; width of the wrapped box
  \centering
  \vspace{-2.1\baselineskip} % users can freely adjust top position by changing this
  \includegraphics[width=\linewidth]{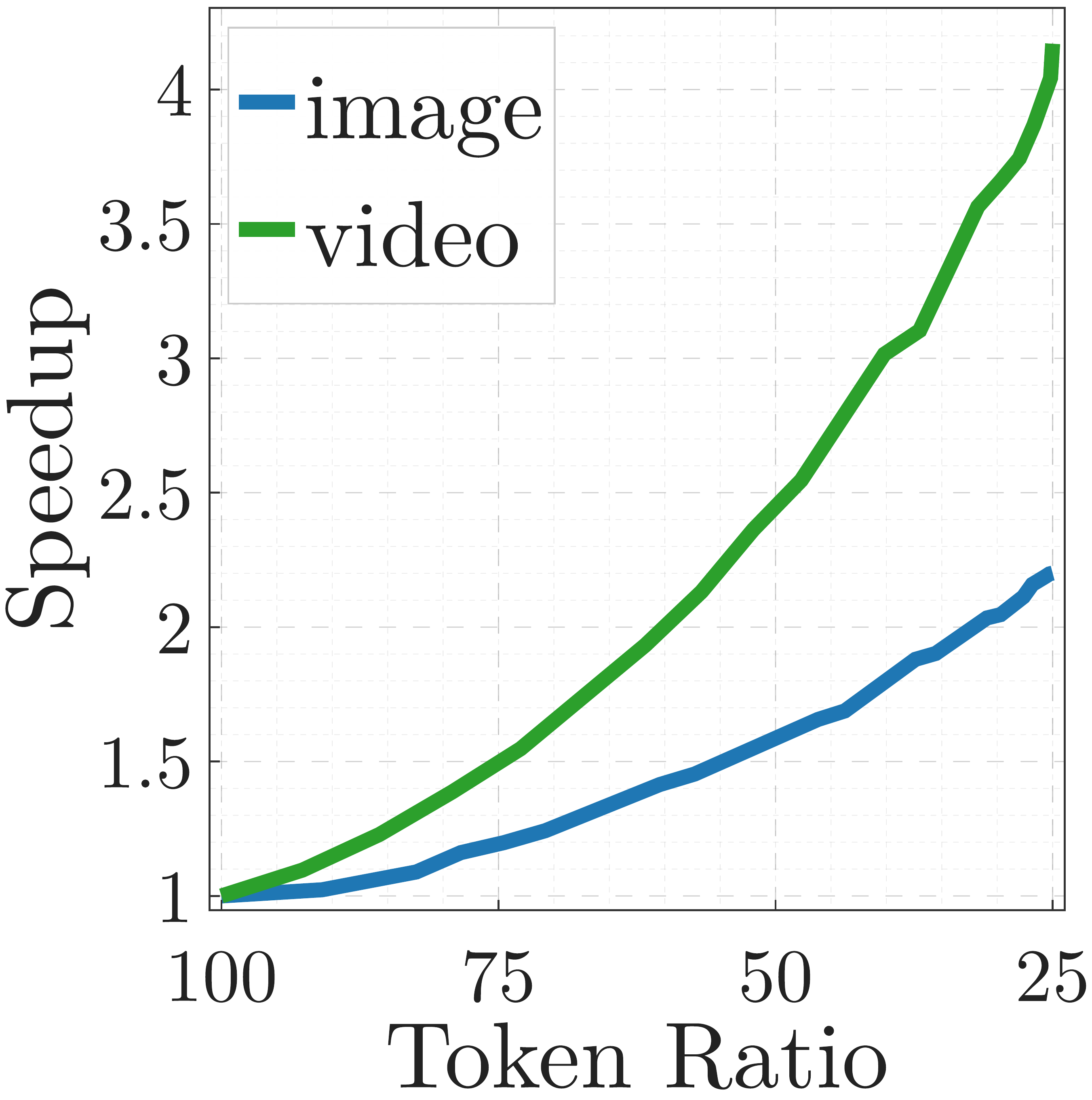}
  \vspace{-22pt}
  \caption{\textbf{Foveated visual generation speedup.}}
  \label{fig:runtime}
  \vspace{-2.0\baselineskip} % (optional) tighten space below
\end{wrapfigure}

\subsubsection{Runtime Comparison.} 
Foveated Diffusion greatly reduces computational complexity via foveated tokenization, significantly accelerating visual generation. This computational efficiency is determined by the foveation mask (see Sec. \ref{subsubsc:fov_token}).
By expanding the low-resolution periphery, we drastically reduce the effective sequence length to $L$ tokens, compared to $h\!\cdot\!w$ for images and $f\!\cdot\!h\!\cdot\!w$ for videos.
We define Token Ratio as the proportion of the reduced sequence relative to the full sequence and measure the resulting computational savings compared to full high-resolution generation as Speedup.
As shown in Fig.~\ref{fig:runtime}, using $25\%$ of the tokens yields remarkably over $2\times$ and over $4\times$ speedup for image and video generation, respectively. Video models achieve higher speedup due to the higher computational cost of spatiotemporal (3D) attention operations.

\input{tables/image_gen}

\subsubsection{Image Generation.} 
\label{subsubsec:image_gen}
For foveated image generation, we report standard generative metrics including FID \cite{heusel2017fid}, Precision \cite{kynkaanniemi2019precision}, and a human preference metric HPSv2.1 \cite{wu2023hps}, and measure prompt alignment using CLIP score \cite{radford2021clip}. For the na\"ive mixed-resolution baseline and our Foveated Diffusion pipeline, we fix the foveation mask to be a centered rectangular mask with varying token ratios.

As shown in Tab. ~\ref{tab:image_metrics} and Fig. \ref{fig:baseline_image}, Foveated Diffusion consistently achieves substantially better performance than the na\"ive mixed-resolution baseline across all foveation sizes and across all metrics aside from FID, while maintaining performance similar to full high-resolution generation. Importantly, our method significantly surpasses the na\"ive baseline on a human preference–aligned metric HPSv2.1 \cite{wu2023hps}, reinforcing the perceptual, human-centered focus of our approach. We observe that FID may not be a reliable metric for evaluating foveated visual generation, as the naïve baseline, despite exhibiting clear structural artifacts (see Fig. \ref{fig:baseline_image}), even outperforms full high-resolution generation. As the foveation size decreases, the peripheral low-resolution area increases, leading to a significant reduction in generation time.

\begin{wrapfigure}{r}{0.38\textwidth} % r/l for right/left; width of the wrapped box
  \centering
  \vspace{-2.0\baselineskip} % users can freely adjust top position by changing this
  \includegraphics[width=\linewidth]{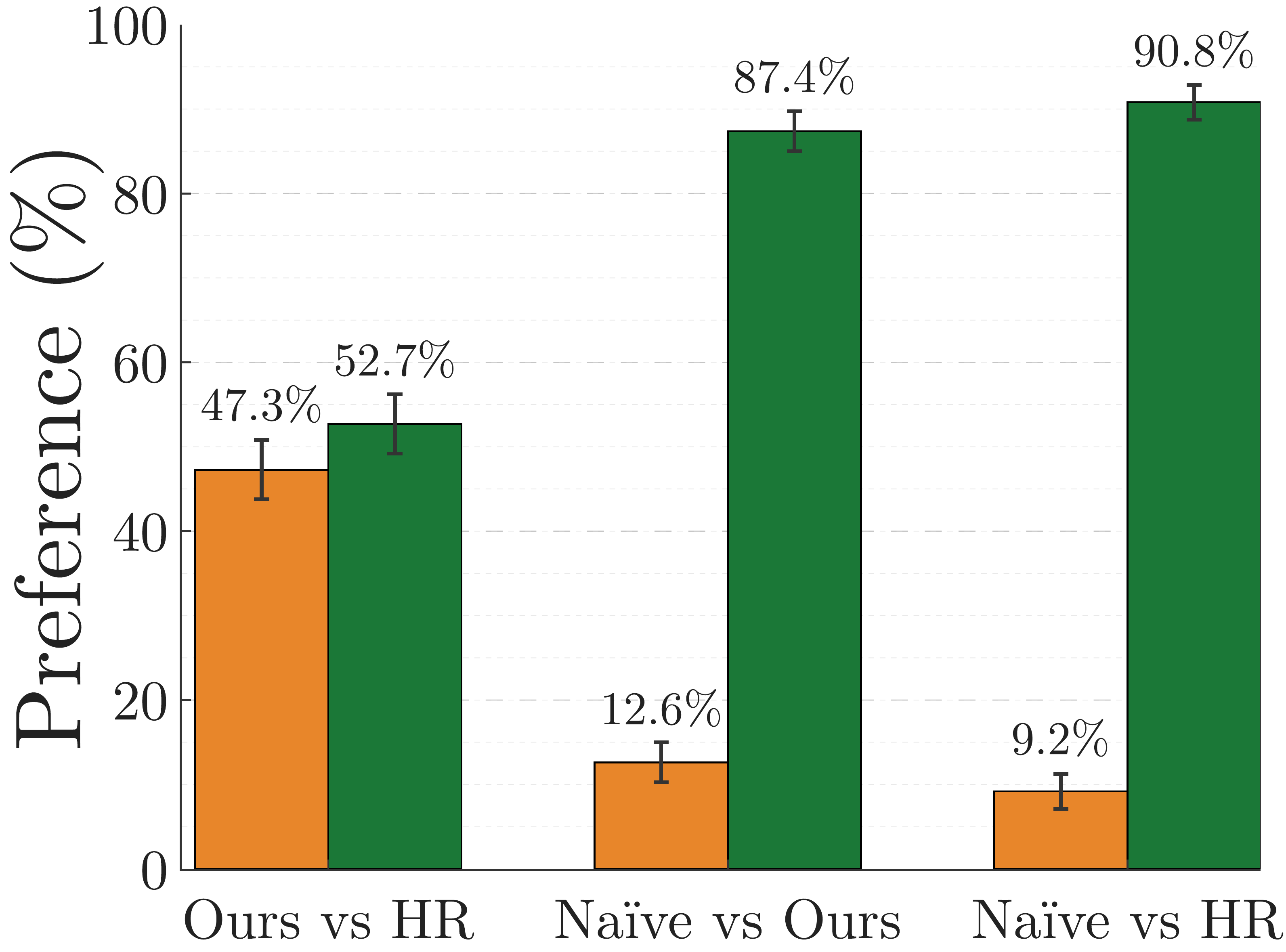}
  \vspace{-20pt}
  \caption{\textbf{User study results.}}
  \label{fig:user_study}
  \vspace{-2.0\baselineskip} % (optional) tighten space below
\end{wrapfigure}

\paragraph{Perceptual User Study.} Standard generative metrics weigh all pixels uniformly and ignore eccentricity-dependent visual acuity, leading to trends that conflict with human preference (e.g., the HPSv2.1–FID discrepancy in Tab. ~\ref{tab:image_metrics}). This is fundamentally misaligned with the perceptually-driven and gaze-contingent motivation of Foveated Diffusion. We therefore perform a Two-Alternative Forced Choice (2AFC) user study under a pseudo-eye-tracked protocol. Participants fixate on a red point before a pair of images randomly drawn from two of the three methods (our method, full high-resolution generation, and the naïve mixed-resolution baseline) are sequentially displayed for $1$ second to discourage eye movements. Participants then select the image with higher overall visual quality, avoiding bias from actively searching for distortions.

In Fig.~\ref{fig:user_study}, we show that our method achieves near perceptual parity with full high-resolution generation and is strongly preferred over the naïve baseline. These results confirm that Foveated Diffusion preserves perceptual quality under gaze-contingent viewing while substantially reducing latency (user study images are generated with a $1.85\times$ speedup), establishing its practicality for real-time, wide-field-of-view applications such as gaming and immersive video. We include a detailed description of our user study design, procedure, analysis, and results in the supplementary materials.

% We use p-value to confirm these results are statistically significant. 
% A two-sided binomial test under the null hypothesis of equal preference yields p-values of $0.48$ (ours vs.\ full high-resolution) and $0.00$ (naïve baseline vs.\ ours); the former ($p > 0.05$) indicates a failure to reject the null hypothesis, confirming statistical indistinguishability from full high-resolution generation, while the latter confirms a significant preference for our method over the naïve baseline. fog

\input{tables/vbench}

\begin{figure}[!t]
    \centering
    \includegraphics[width=\textwidth]{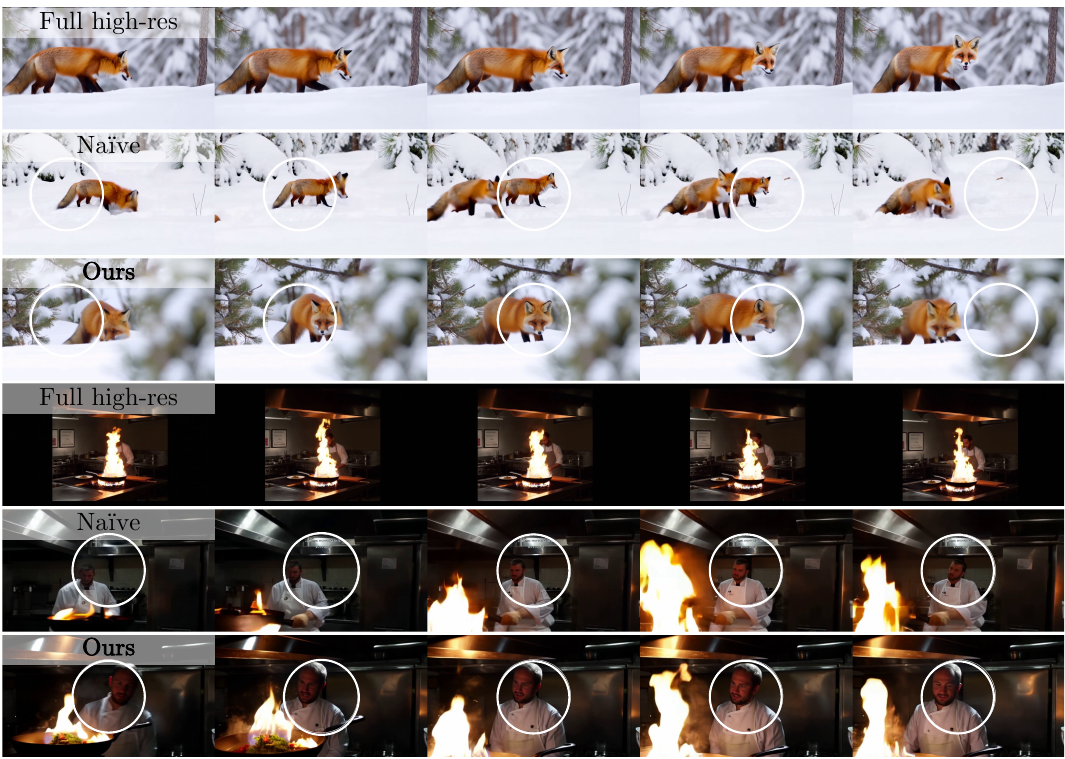}\vspace{-8pt}
    \caption{\textbf{Qualitative comparison for video generation.} 
    Foveated Diffusion outperforms the na\"ive mixed-resolution baseline in video generation, which exhibits scale mismatches or duplicate entities near the low–high resolution boundary (white outline).}
    \vspace{-10pt}
    \label{fig:baseline_video}
\end{figure}

\subsubsection{Video Generation.} 
\label{subsubsec:video_gen}
For foveated video generation, we report the standard generative video metric VBench \cite{huang2023vbench}. Similar to our image generation experiments, we fix the foveation mask across all frames as a centered circular mask with a token ratio of $38 \%$ relative to the original token length of $(f \cdot h \cdot w)$.
Foveated Diffusion surpasses the na\"ive mixed-resolution baseline while achieving results comparable to full high-resolution generation with a $3.5\times$ speedup, as clearly shown in Tab. \ref{tab:video_metrics}. This parity across quality and consistency metrics highlights our framework's ability to maintain a coherent global structure. Fig. \ref{fig:baseline_video} provides supporting qualitative results, where the na\"ive baseline exhibits severe structural and scale mismatches.

\subsubsection{Foveation Mask Strategies.} 
\label{subsubsec:saliency} 
\begin{figure}[!t]
    \centering
    \includegraphics[width=\textwidth]{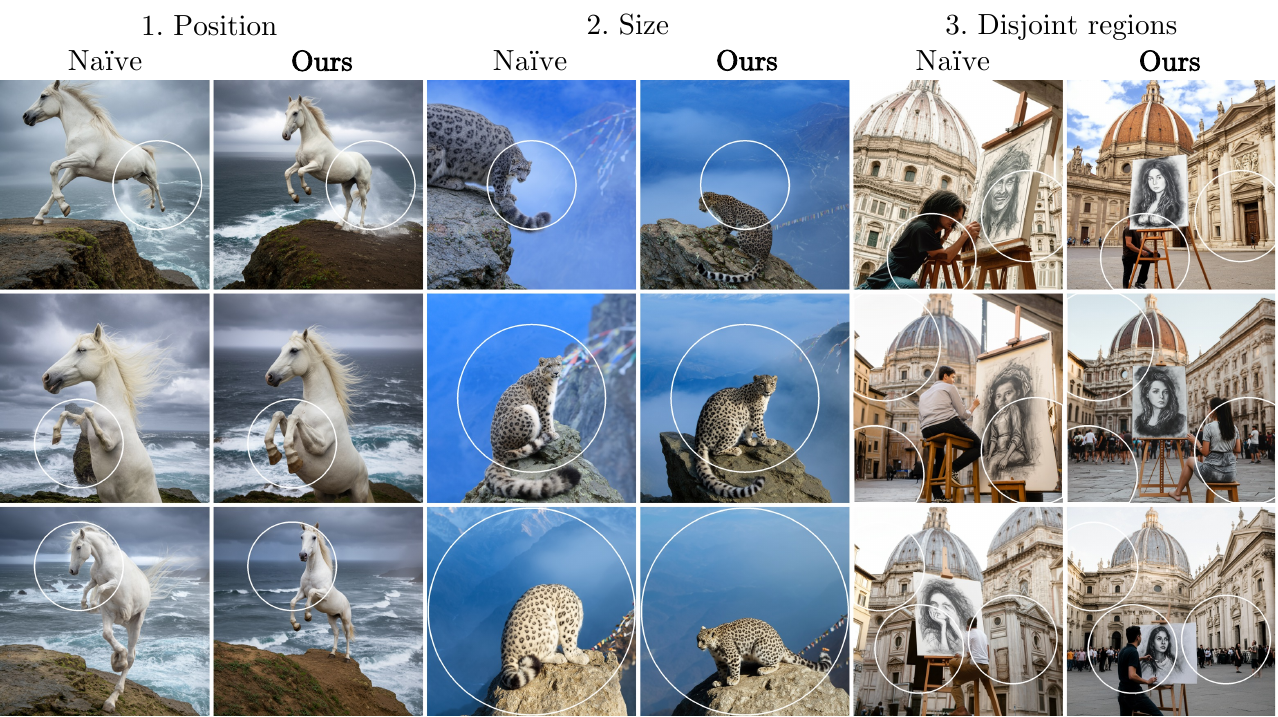}
    \vspace{-17pt}
    \caption{\textbf{Image generation with different foveation patterns. } We generate images using the same prompt and noise seed while varying the foveation pattern in shape, position, and size. High-resolution regions are delineated with white borders. Our method maintains content consistency across resolution regions, while the naïve mixed-resolution baseline exhibits inconsistencies of scale and structure.}
    \label{fig:fov_pattern_image}
\end{figure}

\vspace{-20pt}
As discussed in Sec.~\ref{subsub:foveation_mask_stragegy}, the strategy used to construct foveation masks during training significantly affects the behavior of the resulting foveated generative model.  In Fig. \ref{fig:fov_pattern_image}, we present foveated image generation results using various masks while keeping the text prompt and noise seed constant. We vary the masks in size, position, and shape, including non-contiguous masks with multiple disjoint high-resolution regions. Foveated Diffusion generates coherent content under unseen foveation masks at inference, which is uniquely enabled by our \textit{randomized mask} training protocol.

\begin{figure}[!t]
    \centering
    \includegraphics[width=\textwidth]{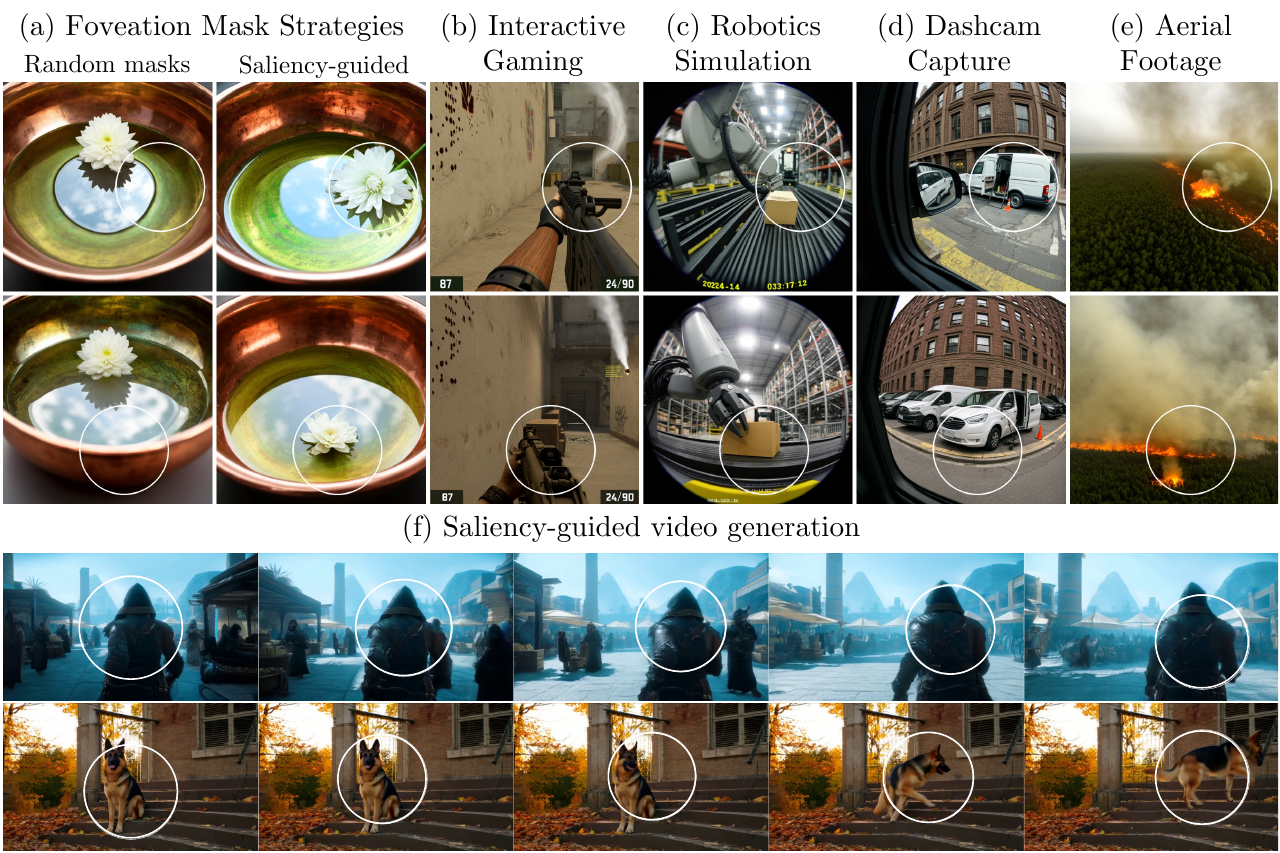}\vspace{-8pt}
    \caption{\textbf{Towards saliency-guided visual generation.} We show that the \textit{saliency-guided} Foveated Diffusion models enable saliency-guided image (a-e) and video generation (f), where salient objects align with the fovea. The \textit{randomized masks} model do not exhibit such behavior (a). This is potentially useful for generative simulation applications such as VR gaming or robotics simulations (b-e), where only the salient objects have to be generated at the highest resolution, i.e. the machine gun in (b),  the robotics arm and box in (c), and the dog in (f).}
    \vspace{-10pt}
    \label{fig:saliency}
\end{figure}

Furthermore,
% by simply replacing randomized foveation masks with binarized saliency maps during training, 
Foveated Diffusion shows great potential for \textit{saliency-guided} generation. Specifically, we construct foveation masks by binarizing image and video saliency maps predicted by DeepGaze \cite{linardos2021deepgaze}.
As evident in Fig.~\ref{fig:saliency}, we observe that salient objects align with specified foveal regions, demonstrating the generalization of our Foveated Diffusion framework beyond random mask placements. This is potentially useful for generative VR gaming or generative robotics simulation scenarios where only salient objects in view are required to be rendered in high resolution \cite{kerrj2025eyerobot}.

\iffalse
\subsubsection{\textbf{Visual Generation with Different Foveation Patterns.}} 
\label{subsubsec:foveation_patterns}
We present visual generation results using different foveation masks $\foveationmask$ while fixing the prompt and noise seed. We vary foveation masks in size, position, and shape, including non-contiguous masks with multiple disjoint high-resolution regions in Figures \ref{fig:fov_pattern_image} and \ref{fig:fov_pattern_video}. Although trained on a discrete set of randomly sampled masks, Foveated Diffusion generalizes to a broad range of unseen foveation patterns for both image and video generation. 
\fi

% \begin{figure}
%     \centering
%     \includegraphics[width=\textwidth]{}
%     \caption{\textbf{Video generation with different foveation patterns. }}
%     \label{fig:fov_pattern_video}
% \end{figure}

%\paragraph{\textbf{LoRA rank and full finetuning.}}

% \paragraph{\textbf{Randomized mask training.}} We investigate the effectiveness of Foveated Diffusion training with randomized foveation patterns and its generality to unseen foveation patterns during inference. We train a Foveated Diffusion image model with only a center foveation mask with a radius ratio of $0.5$ defined with respect to the half of the image diagonal. 

% As shown in Fig.~X, Foveated Diffusion trained with randomized foveation masks generalizes naturally to unseen foveation patterns at inference. In contrast, models trained with a fixed mask exhibit significant structural inconsistencies and cross-scale mismatches across foveated regions.

%% file: tables/image_gen.tex
\definecolor{lightred}{RGB}{255, 204, 204}

{\renewcommand{\arraystretch}{1.0}
\setlength{\aboverulesep}{0.2ex}
\begin{table*}[!t]
  \centering
  \scriptsize
\begin{tabular*}{\textwidth}{@{\extracolsep{\fill}} l cccccc @{}}
    \toprule
    \makecell{Method} &
    \makecell{Token Ratio} &
    \makecell{HPSv2.1 $\uparrow$} &
    \makecell{FID $\downarrow$} &
    \makecell{Precision $\uparrow$} &
    %\makecell{Recall $\uparrow$} &
    \makecell{CLIP $\uparrow$} &
    \makecell{Runtime $\downarrow$ (Speedup $\uparrow$)} \\
\midrule
\noalign{\vskip -3.5pt}
\midrule
    Full high-res & 100$\%$ & 0.280 & 11.38 & 0.792 & 0.292 & 10.45s \\
    \midrule
    Na\"ive mixed-res & \multirow{2}{*}{43$\%$} & 0.268 & \textbf{10.99} & 0.769 & 0.292 & \multirow{2}{*}{6.53s (1.61$\times$)} \\
    \textbf{Ours}      &                         & \textbf{0.279} & 11.38 & \textbf{0.777} & \textbf{0.294 }& \\
    \noalign{\vskip 2pt \hbox to \textwidth{\leaders\hbox to 4pt{\hss\rule{2pt}{0.1pt}\hss}\hfil} \vskip 2pt}
    Na\"ive mixed-res & \multirow{2}{*}{30$\%$} & 0.270 & \textbf{11.70} & 0.762 & 0.292 & \multirow{2}{*}{5.27s (1.98$\times$)} \\
    \textbf{Ours}      &                         & \textbf{0.279} & 11.91 & \textbf{0.789} & \textbf{0.293} & \\
    \noalign{\vskip 2pt \hbox to \textwidth{\leaders\hbox to 4pt{\hss\rule{2pt}{0.1pt}\hss}\hfil} \vskip 2pt}
    Na\"ive mixed-res & \multirow{2}{*}{26$\%$} & 0.275 & 12.83 & 0.775 & 0.292 & \multirow{2}{*}{5.02s (2.08$\times$)} \\
    \textbf{Ours}      &                         & \textbf{0.280} & \textbf{12.62} & \textbf{0.792} & \textbf{0.293} & \\
    \bottomrule
  \end{tabular*}
  \vspace{3pt}
  \caption{\textbf{Quantitative comparison for image generation.} We compare against the naïve mixed-resolution baseline across various token count ratios, highlighting the best result for each. Foveated Diffusion surpasses the baseline, excluding FID which we find less reliable for our task. Our method matches the image quality of full high-resolution generation (top row) while achieving up to a $2\times$ speedup.}
  \label{tab:image_metrics}
\end{table*}
}

%% file: tables/vbench.tex
{\renewcommand{\arraystretch}{1.0}
\setlength{\aboverulesep}{0.2ex}
\begin{table*}[!t]
  \centering
  \scriptsize
  \begin{tabular*}{\textwidth}{@{\extracolsep{\fill}} l cccccc @{}}
    \toprule
    \makecell{Method} &
    \makecell{Subject\\Consistency $\uparrow$} &
    \makecell{Background\\Consistency $\uparrow$} &
    \makecell{Motion\\Smoothness $\uparrow$} &
    \makecell{Dynamic\\Degree $\uparrow$} &
    \makecell{Aesthetic\\Quality $\uparrow$} &
    \makecell{Image\\Quality $\uparrow$} \\
    \midrule
    \noalign{\vskip -3.5pt}
    \midrule
    Full high-res & 0.9407 & 0.9363 & 0.9943 & 0.263 & 0.5434 & 0.653 \\
    \midrule
    Na\"ive mixed-res & 0.9072 & 0.9239 & 0.9899 & \textbf{0.465} & 0.4795 & 0.522 \\
    \textbf{Ours} & \textbf{0.9446} & \textbf{0.9393} & \textbf{0.9946} & 0.265 & \textbf{0.5432} & \textbf{0.587} \\
    \bottomrule
  \end{tabular*}
  \vspace{3pt}
  \caption{\textbf{Quantitative comparison for video generation (VBench).} Foveated Diffusion outperforms the na\"ive mixed-resolution baseline across key metrics, achieving performance comparable to full-resolution generation. Notably, our framework maintains high subject consistency while providing a $3.5\times$  speedup.}
    \vspace{-10pt}
  \label{tab:video_metrics}
\end{table*}
}

%% file: sections/conclusions.tex
\vspace{-5pt}
In this work, we introduce Foveated Diffusion, a perceptually motivated framework for efficient visual generation. By leveraging the eccentricity-dependent nature of the human visual system, we achieve significant computational savings while maintaining the perceived quality of the generated content.

Our method yields promising results for foveated generation, generating coherent content across low- and high-resolution regions. Nevertheless, we observe occasional color artifacts near the foveation boundary (see the supplementary materials). This is primarily due to the blending of the VAE-decoded low- and high-resolution content (see Eq.~\ref{eq:vae_decode} and~\ref{eq:image_merge}). A promising direction for future work is redesigning the VAE to directly encode and decode mixed-resolution tokens.
Additionally, we present two levels of foveation with a spatial reduction factor of $2\times 2$, whereas traditional foveated rendering can employ even coarser peripheral resolutions. Our general framework naturally extends to multiple levels of foveation, and such an extension calls for the corresponding multi-level adaptation of phase-aligned RoPE \cite{wu2025crpa}.
Finally, we believe that our method is most impactful when deployed on a streaming autoregressive video generation system equipped with an eye tracker. While we are the first to establish the foundations of such a system, integrating Foveated Diffusion into a real-time video world model remains a compelling direction for future work.

In conclusion, Foveated Diffusion offers a new paradigm and opens a new avenue for scaling generative models: aligning model computation with human visual perception, complementary to advances in hardware and algorithmic efficiency.

%% file: supp_sections/implementation.tex
\subsection{Image Generation}
\label{supp_subsec:image_impl}
For our foveated image generation experiments, we fine-tune the FLUX 2.1 Klein 4B model \cite{blackforestlabs2025flux2klein} using the Aesthetic-Train-V2 dataset \cite{zhang2025diffusion4k, zhang2025ultrahighresolutionimagesynthesis}. We adopt a Low-Rank Adaptation (LoRA) \cite{hu2022lora} approach with a rank of 32, training for 10{,}000 steps. The optimization was conducted on a cluster of eight NVIDIA H100 GPUs with an effective batch size of 8. Our implementation utilizes the DiffSynth-Studio codebase \footnote{\url{https://github.com/modelscope/DiffSynth-Studio/tree/main}} and follows its default hyperparameter settings for LoRA training.

For quantitative results, we generate 10K images, one for each prompt in the reserved test set from the Aesthetic-Train-V2 dataset~\cite{zhang2025diffusion4k, zhang2025ultrahighresolutionimagesynthesis}. We adopt the standard evaluation protocol in~\cite{dhariwal2021diffusion} and report standard generative metrics including HPSv2.1~\cite{wu2023hps}, FID~\cite{heusel2017fid}, Precision~\cite{kynkaanniemi2019precision}, and CLIP score~\cite{radford2021clip} in Table~1 of the main paper. FID and Precision are computed against real images in the evaluation set, reflecting the data alignment between generated and real images. The CLIP and HPSv2.1 scores are averaged across all generated images, where the CLIP score measures prompt alignment and the HPSv2.1 score captures human preference.

For all image generation experiments, we generate images at $1024\times1024$ resolution for all methods.

\subsection{Video Generation}
\label{supp_subsec:video_impl}
For our foveated video generation experiments, we fine-tune the Wan2.1 1.3B model \cite{wan2025wan} using the Vchitect-T2V-Dataverse dataset \cite{fan2025vchitect,si2025RepVideo}. We adopt a Low-Rank Adaptation (LoRA) \cite{hu2022lora} approach with a rank of 32, training for 10{,}000 steps. The optimization was conducted on a cluster of eight NVIDIA H100 GPUs with an effective batch size of 8. Our implementation utilizes the DiffSynth-Studio codebase and follows its default hyperparameter settings for LoRA training.

For quantitative evaluations, we generate $200$ videos using the held-out test prompts and report the standard VBench \cite{huang2023vbench} metrics. 
We evaluate our approach and all baselines at a consistent 480p resolution.

For qualitative results, we provide samples at the original 480p training resolution and additionally demonstrate generalization to 720p.

%% file: supp_sections/user_study.tex
\subsection{Study Design and Participants}
\label{supp_subsec:user_study_design}
\noindent\textbf{Study design.} We evaluate the perceptual quality of Foveated Diffusion against both full high-resolution generation and the naïve mixed-resolution baseline using a Two-Alternative Forced Choice (2AFC) paradigm, a standard protocol for preference-based perceptual evaluation~\cite{krajancich2023towards}. In each trial, participants are shown a pair of images sequentially and are asked to select the one they judge to have higher overall visual quality. Forced choice eliminates neutral or indecisive responses and yields a clean preference rate $P \in [0, 1]$, where the null hypothesis of perceptual equivalence corresponds to $P = 0.5$.

Since Foveated Diffusion targets gaze-contingent generation as a primary use case, the study employs a pseudo–eye-tracking protocol. Rather than using physical eye-tracking hardware, each trial begins with a red fixation dot displayed on a black screen at the position corresponding to the center of the foveal region for that trial. Participants fixate on this dot before each full-screen image is shown, ensuring that their gaze is directed toward the foveal center. For foveated and naïve mixed-resolution images, the dot is placed precisely at the center of the high-resolution region. Although full high-resolution images have uniform resolution across the entire image, participants fixate at the same location for consistency. Detailed trial procedures are described in Sec.~\ref{supp_subsec:user_study_procedure}.

\noindent\textbf{Study participants.} A total of 11 participants (8 male and 3 female; ages 21--32) took part in the study. All participants reported normal or corrected-to-normal vision, no history of visual deficiencies, and no color blindness. All participants provided informed consent.

\subsection{Study Setup and Images}
\label{supp_subsec:user_study_setup}
\begin{wrapfigure}[11]{r}{0.4\textwidth} % r/l for right/left; width of the wrapped box
  \centering
  \vspace{-2.1\baselineskip} % users can freely adjust top position by changing this
  \includegraphics[width=\linewidth]{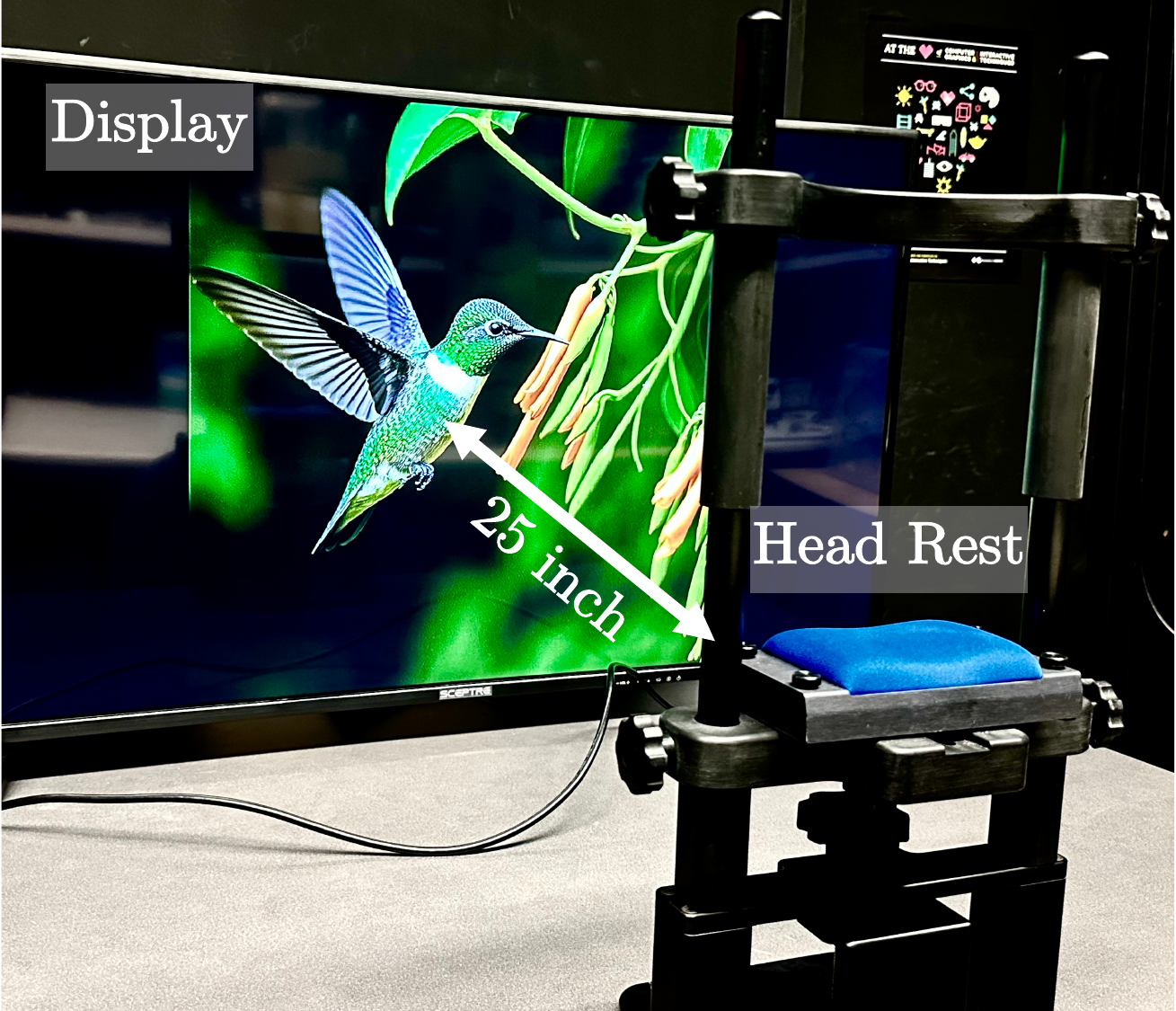}
  \vspace{-22pt}
  \caption{\textbf{User study setup.}}
  \label{fig:user_study_setup}
  \vspace{-2.0\baselineskip} % (optional) tighten space below
\end{wrapfigure}

Due to the need to display high-resolution content, we used a Sceptre X405BV-FSR LED monitor (40-inch, 16:9 aspect ratio) at a native resolution of $1920 \times 1080$ (Full HD), a 60\,Hz refresh rate, and a peak luminance of 250\,cd/m$^2$ for image display. All test images were square and centered on the display to preserve their aspect ratio. The experiment was implemented in Python using the PsychoPy package~\cite{peirce2007psychopy}, and images were streamed to the display via a wired HDMI connection.

Participants were seated at a fixed viewing distance of 25 inches (approximately 63.5\,cm), maintained by a headrest that also controlled viewing height (Fig.~\ref{fig:user_study_setup}). At this distance, the display subtends approximately 24 pixels per degree (ppd) of visual angle, so each full-screen image spans roughly $45^\circ \times 45^\circ$ of the visual field.

The foveal region diameter was set to one-third of the image width, subtending approximately $15^\circ$ of visual angle at the prescribed viewing distance. This boundary was chosen based on the known eccentricity-dependent decline in human visual acuity~\cite{geisler1998foveated}, placing the peripheral region beyond the zone of high acuity. Images were drawn from 40 unique prompts in the same test set used for the quantitative evaluations in Table~1 of the main paper, spanning diverse scenes and objects. For each prompt, foveated and naïve mixed-resolution images were generated with matching foveal region locations, which were randomized across trials (see main paper Sec.~3), ensuring that participants could not anticipate the foveal center from one trial to the next.

\subsection{Procedure}
\label{supp_subsec:user_study_procedure}
\begin{figure}[!t]
    \centering
    \includegraphics[width=1.0\textwidth]{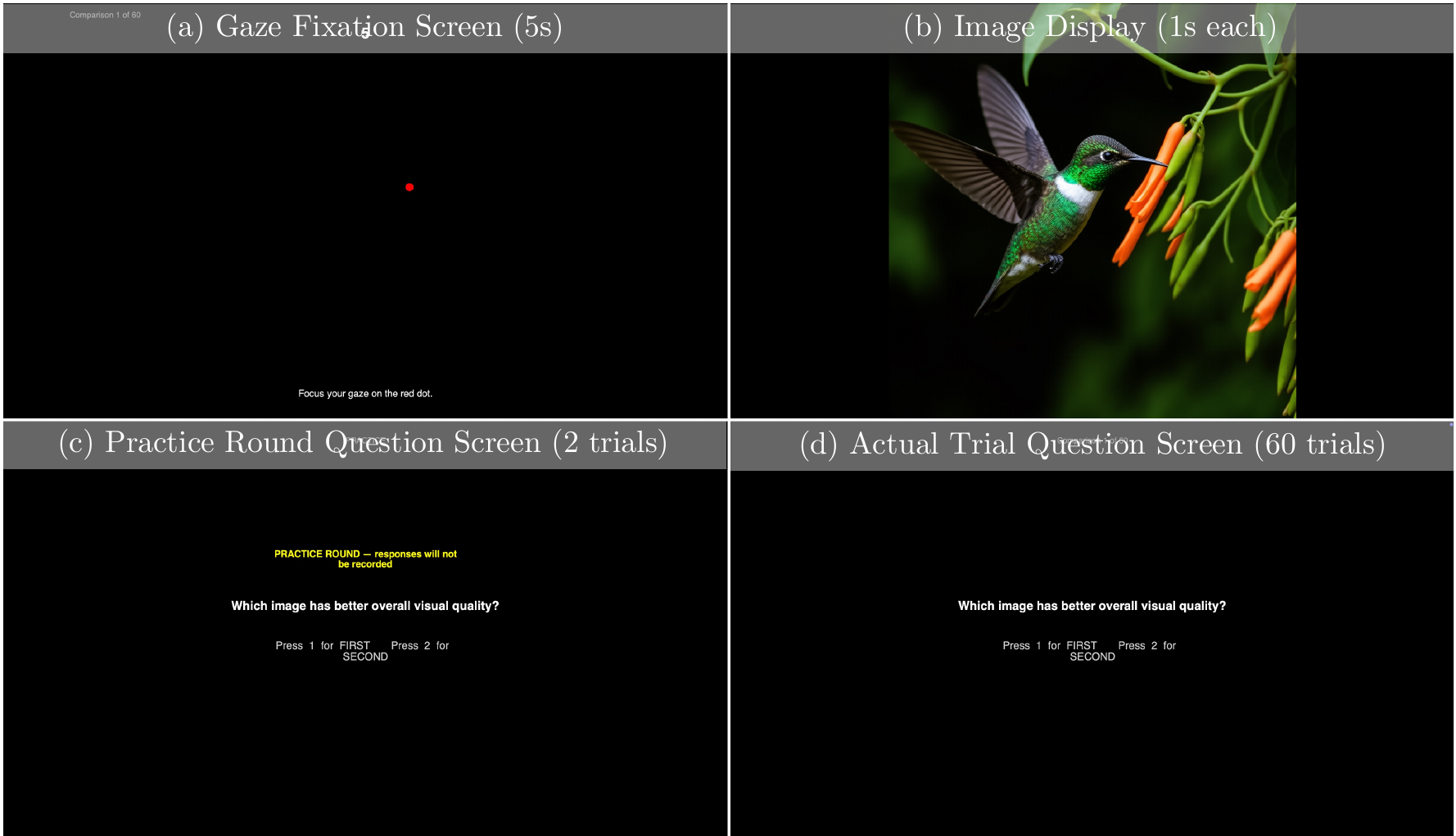}
    \caption{\textbf{User study interface.} Each trial began with a red gaze fixation dot displayed for 5 seconds (\textbf{a}), followed by a pair of images, each displayed for 1\,s (\textbf{b}). Participants answered visual quality questions after both images were shown. Participants completed two practice trials (\textbf{c}) before beginning 60 data collection trials (\textbf{d}).}
    \label{fig:user_study_interface}
\end{figure}

Each participant completed 60 trials, divided equally across three pairwise comparison conditions: Foveated Diffusion vs.\ full high-resolution generation, Foveated Diffusion vs.\ the naïve mixed-resolution baseline, and full high-resolution generation vs.\ the naïve mixed-resolution baseline (20 trials each). Within each condition, the two presentation orders were counterbalanced equally (10 trials per order). For each trial, the test image was randomly selected from the 40 available, with the comparison condition and presentation order assigned according to the counterbalanced scheme.

Each trial began with a five-second red fixation dot at the center of the foveal region, orienting participants' gaze before the first image was shown. The first image was then displayed for one second, followed by a one-second fixation dot to recalibrate gaze, and then the second image for one second. This brief exposure duration limited peripheral exploration, specifically testing whether Foveated Diffusion remained perceptually indistinguishable from full high-resolution generation when participants could perceive little peripheral content. After both images had been shown, participants pressed a keyboard key to indicate which image--the first or the second--had higher overall visual quality (Fig.~\ref{fig:user_study_interface}).

Before data collection, participants completed two practice trials to familiarize themselves with the interface and task. All trials were conducted without breaks; the total completion time was approximately 10 minutes.

\subsection{Statistical Analysis}
\label{supp_subsec:user_study_stats}
\noindent\textbf{Preference rate estimation.}
We group all participants' votes for each pairwise condition, and the preference rate $P$ for each pairwise condition is calculated as the fraction of votes for the target method. Across all participants, each pairwise condition contains 220 data points (votes) in total.

\noindent\textbf{Significance testing.}
To test whether a pairwise preference rate differs significantly from chance, we apply a two-sided binomial test under the null hypothesis $H_0\colon P = 0.5$ (equal preference). A result is considered statistically significant at the $\alpha = 0.05$ level. A $p$-value above $0.05$ for a pair indicates failure to reject $H_0$, i.e., the two methods are perceptually indistinguishable; a $p$-value below $0.05$ indicates a statistically significant preference for one method over the other. Table~\ref{tab:user_study_full} reports the preference rates and $p$-values for all three pairwise conditions.

% \begin{table}[!t]
% \centering
% \setlength{\tabcolsep}{6pt}
% \begin{tabular}{lcc}
% \toprule
% \textbf{Pair} & \textbf{Preference $P$ (\%)} & \textbf{$p$-value ($H_0: P = 0.5$)} \\
% \midrule
% Ours vs.\ High-Res      & 47.3 & 0.4829 \\
% Ours vs.\ Na\"{i}ve     & 87.4 & $<0.0001$ \\
% High-Res vs.\ Na\"{i}ve & 90.8 & $<0.0001$ \\
% \bottomrule
% \end{tabular}
% \caption{\textbf{User study statistical analysis.} Preference rate for the first-listed method in each pair (higher = more preferred). $p$-values are from a two-sided binomial test under $H_0\colon P{=}0.5$. $p > 0.05$ indicates failure to reject $H_0$ and implies perceptual indistinguishability.}
% \label{tab:user_study_full}
% \end{table}
{\renewcommand{\arraystretch}{1.0}
\setlength{\aboverulesep}{0.2ex}
\begin{table*}[!t]
  \centering
  \scriptsize
  \begin{tabular*}{\textwidth}{@{\extracolsep{\fill}} l cc @{}}
    \toprule
    \makecell[l]{Pair} & 
    \makecell{Preference $P$ (\%)} & 
    \makecell{$p$-value ($H_0: P = 0.5$)} \\
    \midrule
    \noalign{\vskip -3.5pt}
    \midrule
    Ours vs.\ Full high-res   & 47.3 & 0.4829 \\
    \textbf{Ours vs.\ Na\"ive mixed-res}  & \textbf{87.4} & \textbf{$<0.0001$} \\
    Full high-res vs.\ Na\"ive mixed-res & 90.8 & $<0.0001$ \\
    \bottomrule
  \end{tabular*}
  \vspace{3pt}
  \caption{\textbf{User study statistical analysis.} Preference rate for the first-listed method in each pair (higher = more preferred). $p$-values are from a two-sided binomial test under $H_0: P=0.5$. $p > 0.05$ indicates failure to reject $H_0$ and implies perceptual indistinguishability.}
  \vspace{-10pt}
  \label{tab:user_study_full}
\end{table*}
}
Based on the $p$-values in Table~\ref{tab:user_study_full}, our user study confirms that Foveated Diffusion is perceptually indistinguishable from full high-resolution generation under gaze-contingent viewing conditions. Both our method and full high-resolution generation are significantly preferred over the na\"{i}ve baseline, due to visual artifacts in the na\"{i}ve baseline generations (main paper Fig.~4, Fig.~5).

%% file: supp_sections/image_qualitative.tex
\subsection{Extended Image Generation Baseline Comparisons}
\label{supp_subsec:image_baseline}
\begin{figure}
    \centering
    \includegraphics[width=1.0\textwidth]{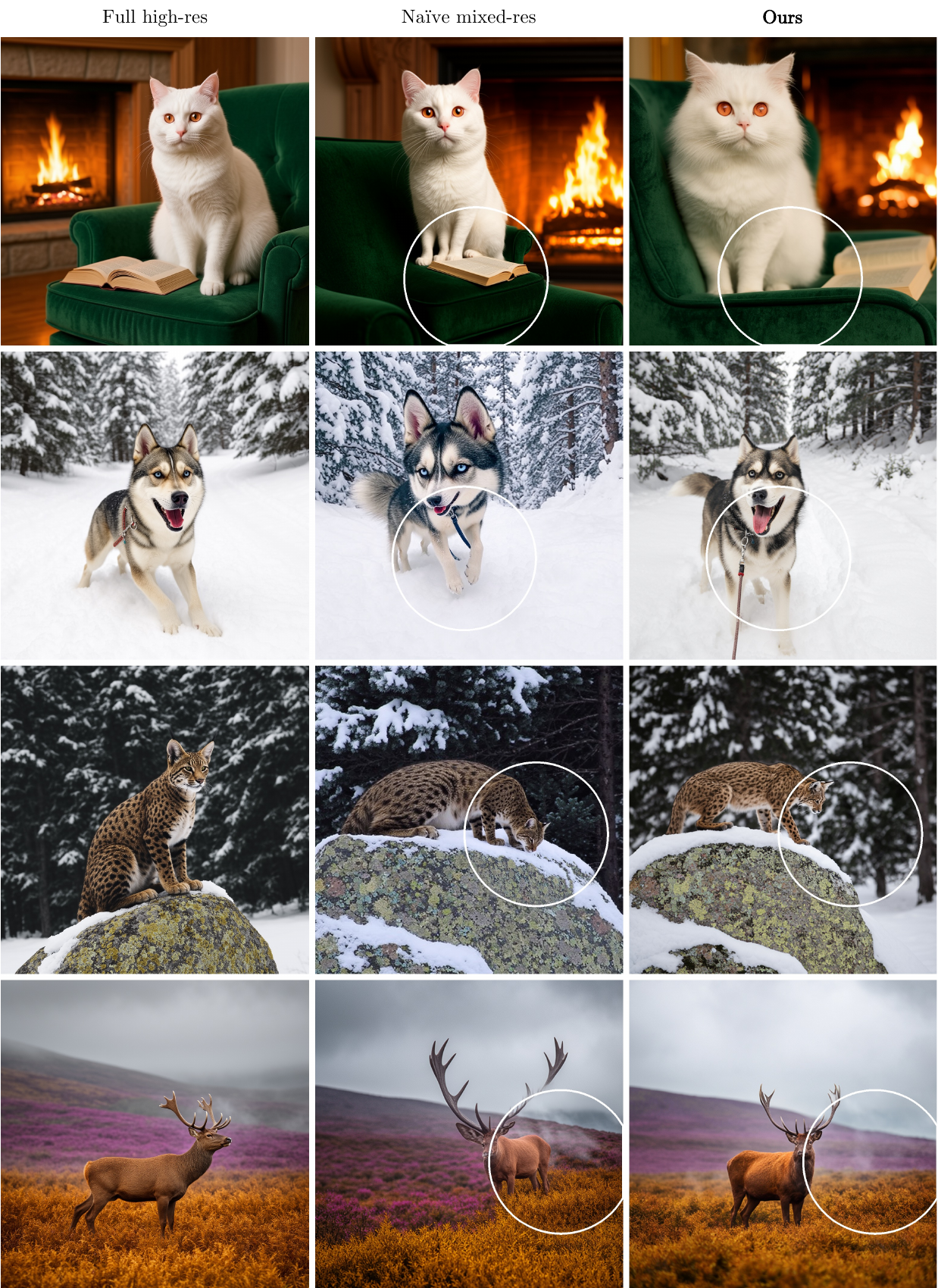}
    \caption{\textbf{Extended baseline comparisons.} Foveated Diffusion (ours) produces perceptually similar quality to full high-resolution generation, whereas the na\"ive mixed-resolution baseline exhibits severe scale mismatches and structural inconsistencies across resolutions. All images are uncompressed in this figure.}
    \label{fig:baseline_1}
\end{figure}

\begin{figure}
    \centering
    \includegraphics[width=0.9\textwidth]{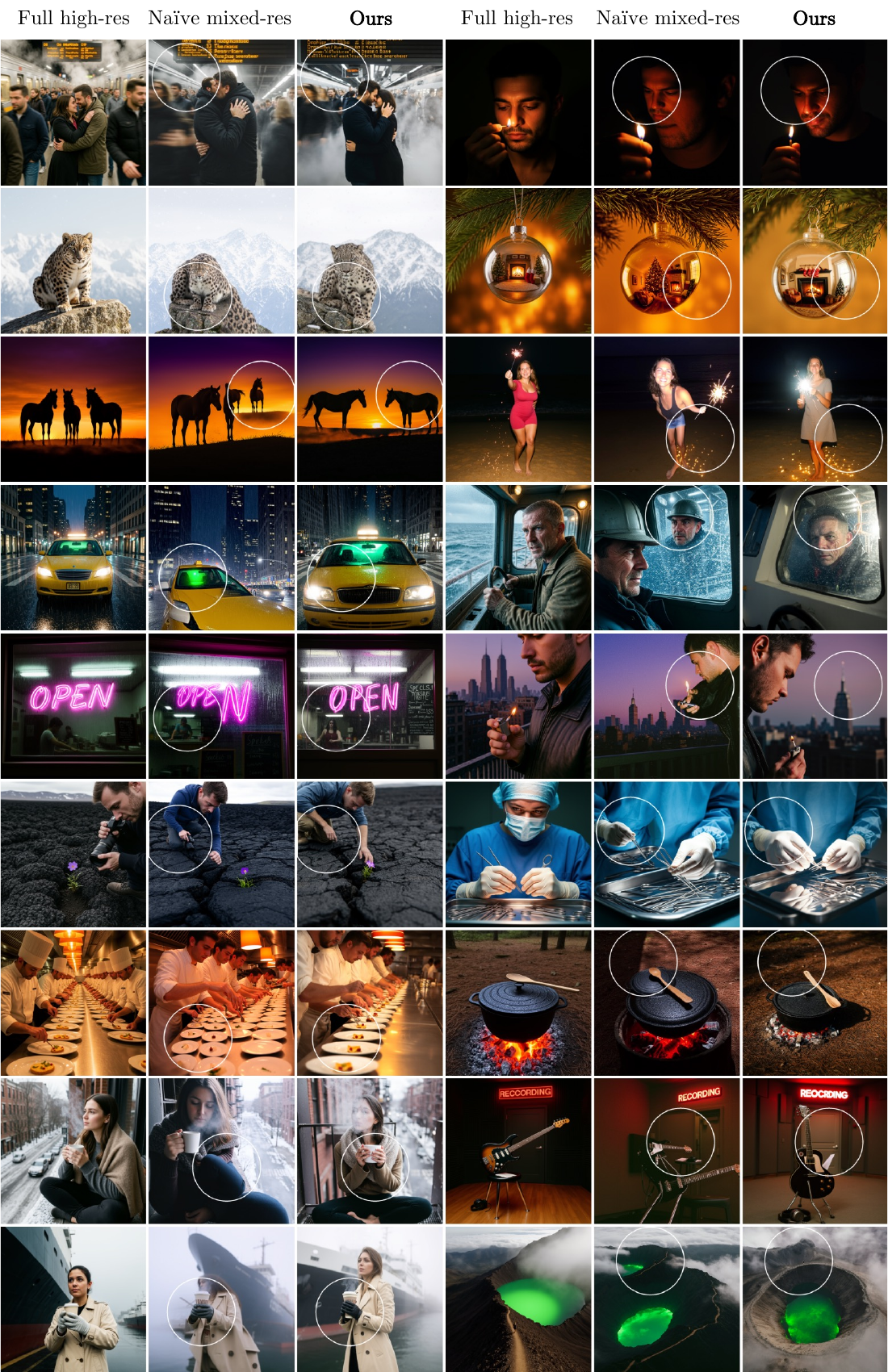}
    \caption{\textbf{Extended baseline comparisons.} Foveated Diffusion (ours) produces perceptually similar quality to full high-resolution generation, whereas the na\"ive mixed-resolution baseline exhibits severe scale mismatches and structural inconsistencies across resolutions.}
    \label{fig:baseline_2}
\end{figure}

\begin{figure}
    \centering
    \includegraphics[width=0.9\textwidth]{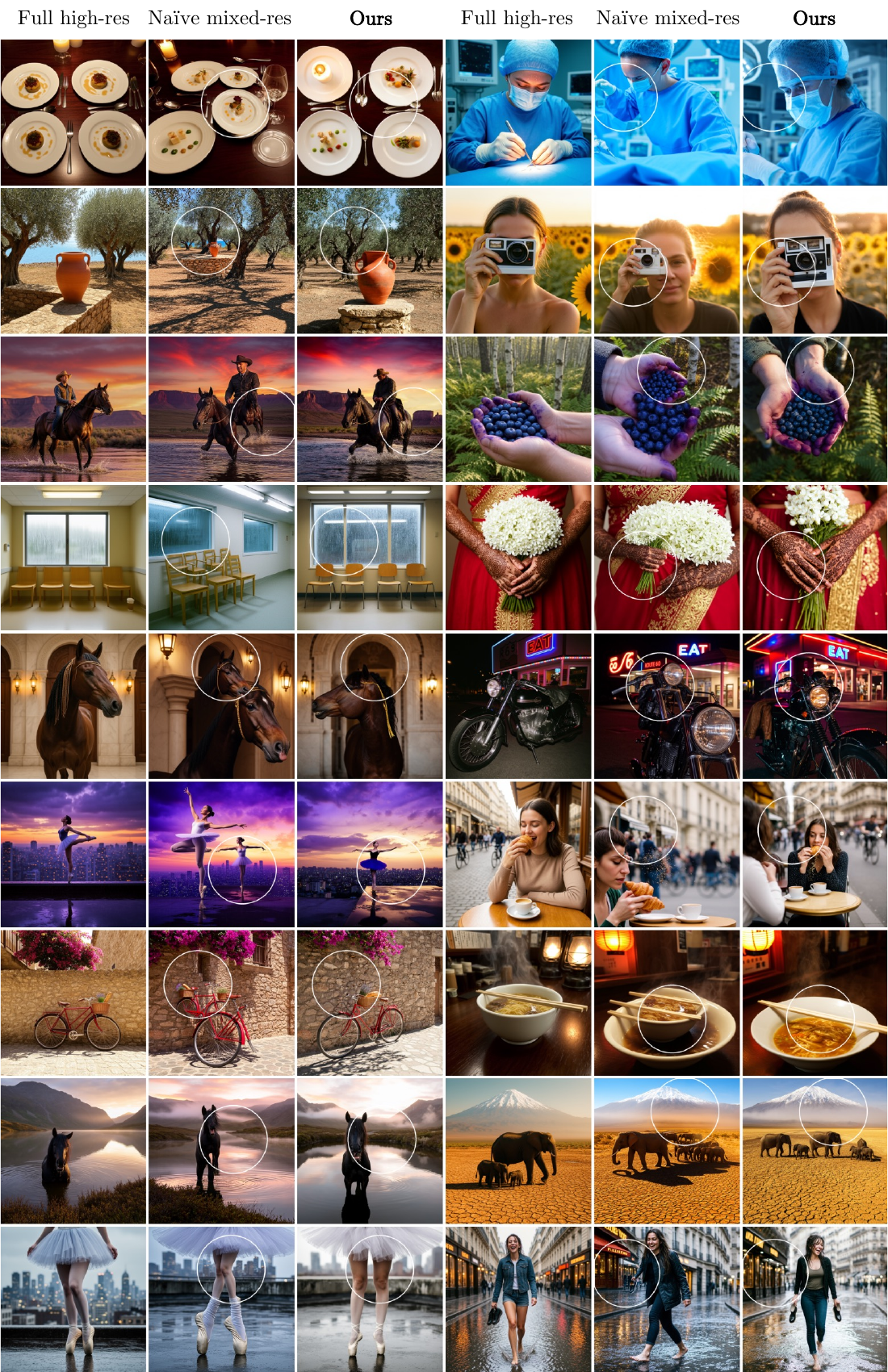}
    \caption{\textbf{Extended baseline comparisons.} Foveated Diffusion (ours) produces perceptually similar quality to full high-resolution generation, whereas the na\"ive mixed-resolution baseline exhibits severe scale mismatches and structural inconsistencies across resolutions.}
    \label{fig:baseline_3}
\end{figure}

\begin{figure}
    \centering
    \includegraphics[width=0.9\textwidth]{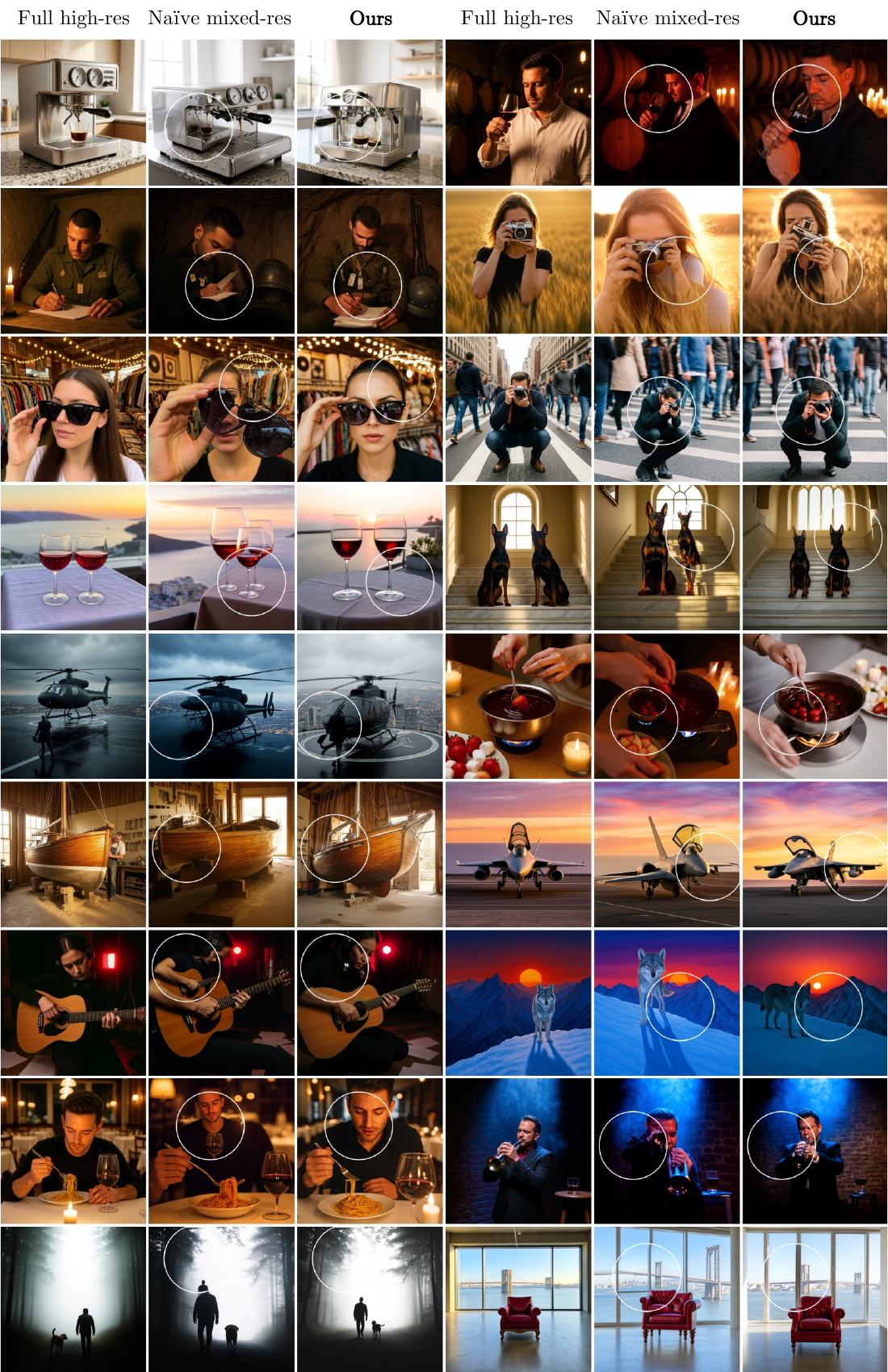}
    \caption{\textbf{Extended baseline comparisons.} Foveated Diffusion (ours) produces perceptually similar quality to full high-resolution generation, whereas the na\"ive mixed-resolution baseline exhibits severe scale mismatches and structural inconsistencies across resolutions.}
    \label{fig:baseline_4}
\end{figure}

In Figures \ref{fig:baseline_1}-\ref{fig:baseline_4}, we provide extended baseline comparisons against full high-resolution generation and na\"ive mixed-resolution generation using our \textit{randomized mask model}. The high-resolution region defined by the foveation mask is a circle with radius $0.5$ relative to the image diagonal and a randomly placed center.

Our method consistently outperforms the na\"ive mixed-resolution baseline, generating coherent and consistent content with consistent structure and scale, whereas the mixed-resolution baseline exhibits significant distortions. Most importantly, our method achieves perceptually indistinguishable quality from the full high-resolution baseline while using approximately $57\%$ fewer tokens, resulting in a $1.85\times$ speedup in image generation time.

\clearpage

\clearpage

\clearpage

\clearpage

\clearpage
\subsection{Image Generation with Different Foveation Patterns} 
\label{supp_subsec:image_fov}

\begin{figure}
    \centering
    \includegraphics[width=0.9\textwidth]{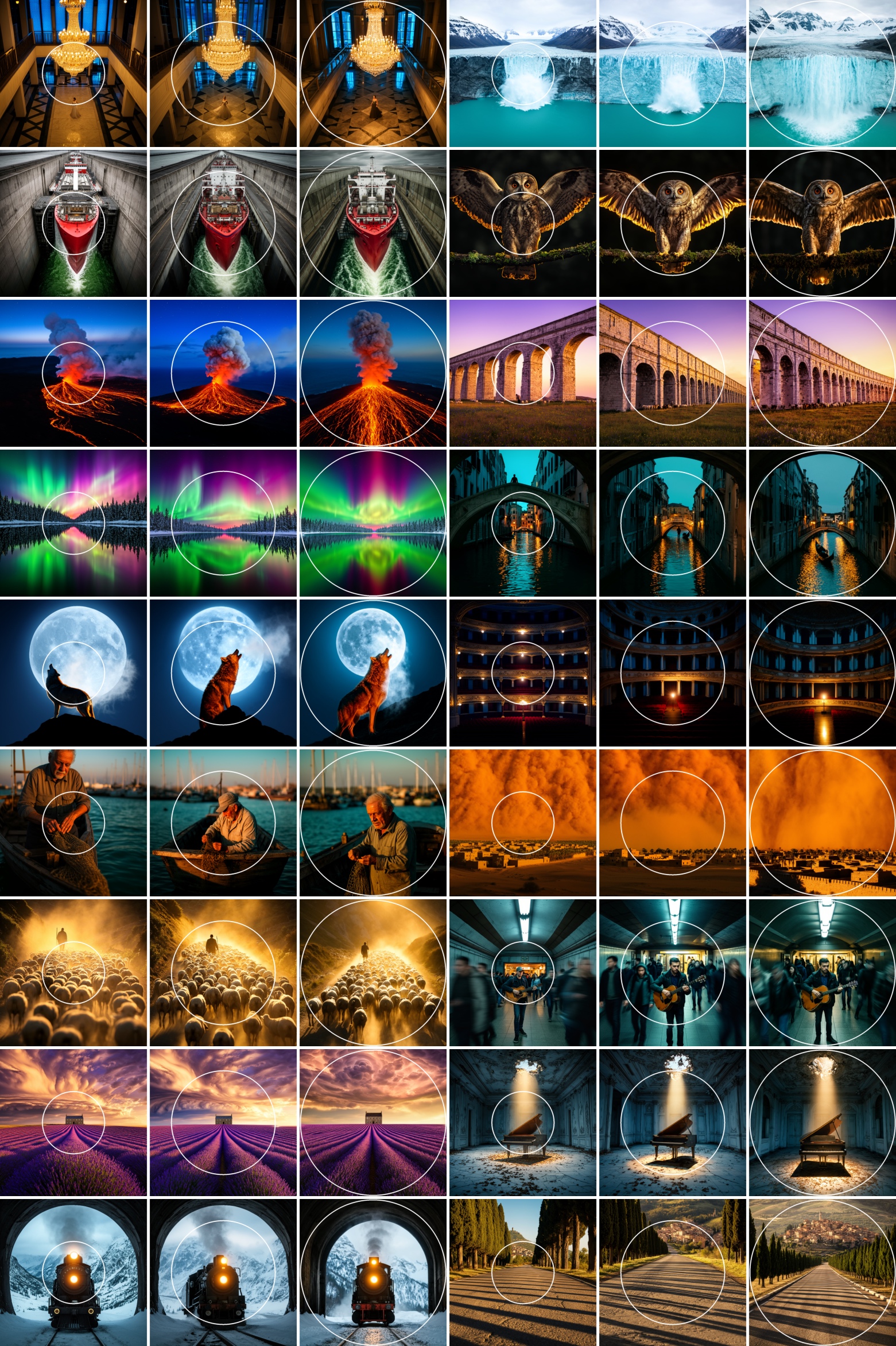}
    \caption{\textbf{Varying foveation mask radius.} Foveated Diffusion generates coherent content across varying foveation mask radii. Given the same prompt and noise seed, the generated images remain consistent and adhere to the prompt.}
    \label{fig:fov_pattern_radius}
\end{figure}

\begin{figure}
    \centering
    \includegraphics[width=0.9\textwidth]{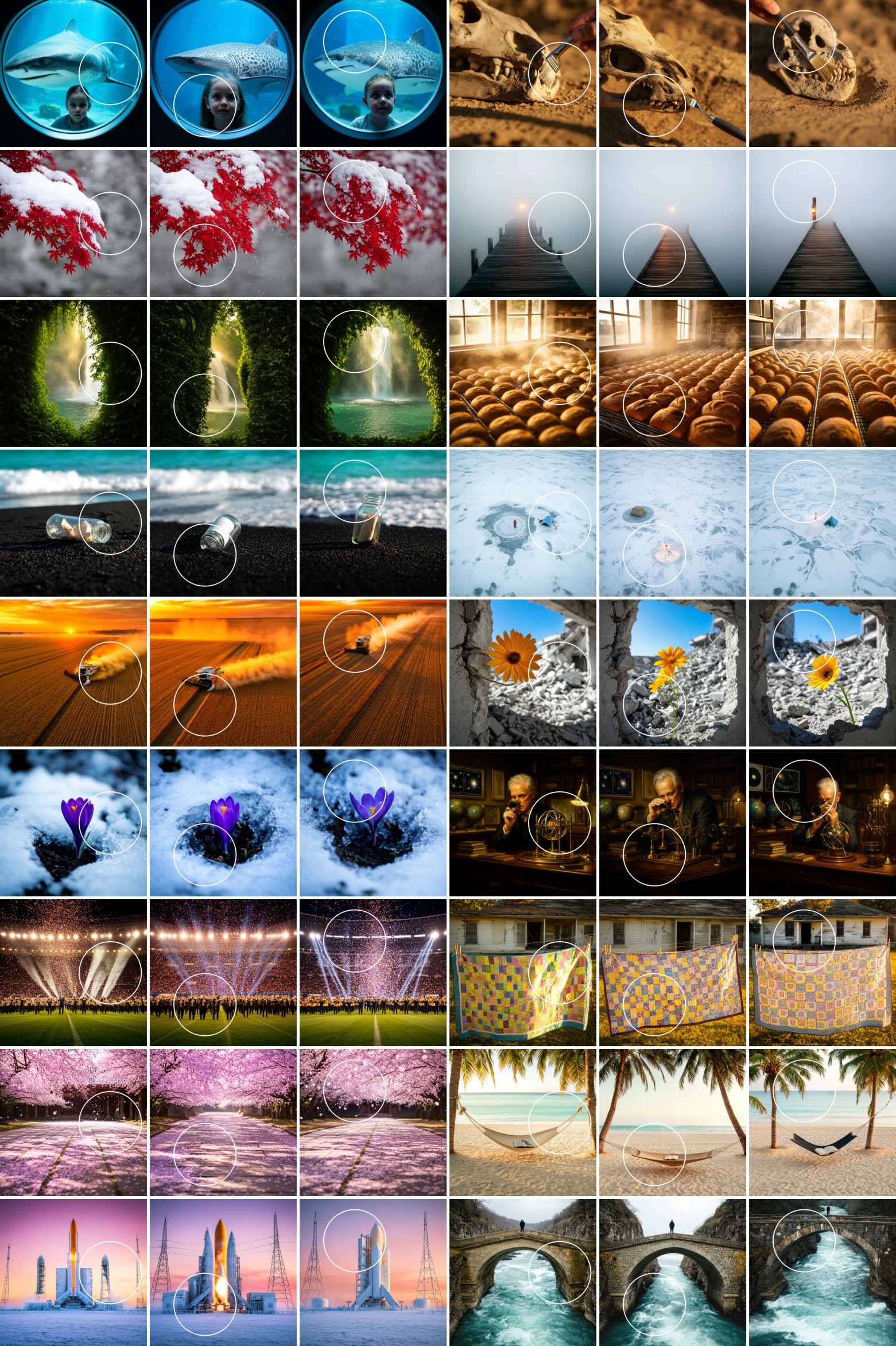}
    \caption{\textbf{Varying foveation mask position.} Foveated Diffusion generates coherent content across varying foveation mask positions. Given the same prompt and noise seed, the generated images remain consistent and adhere to the prompt.}
    \label{fig:fov_pattern_circular}
\end{figure}

\begin{figure}
    \centering
    \includegraphics[width=0.9\textwidth]{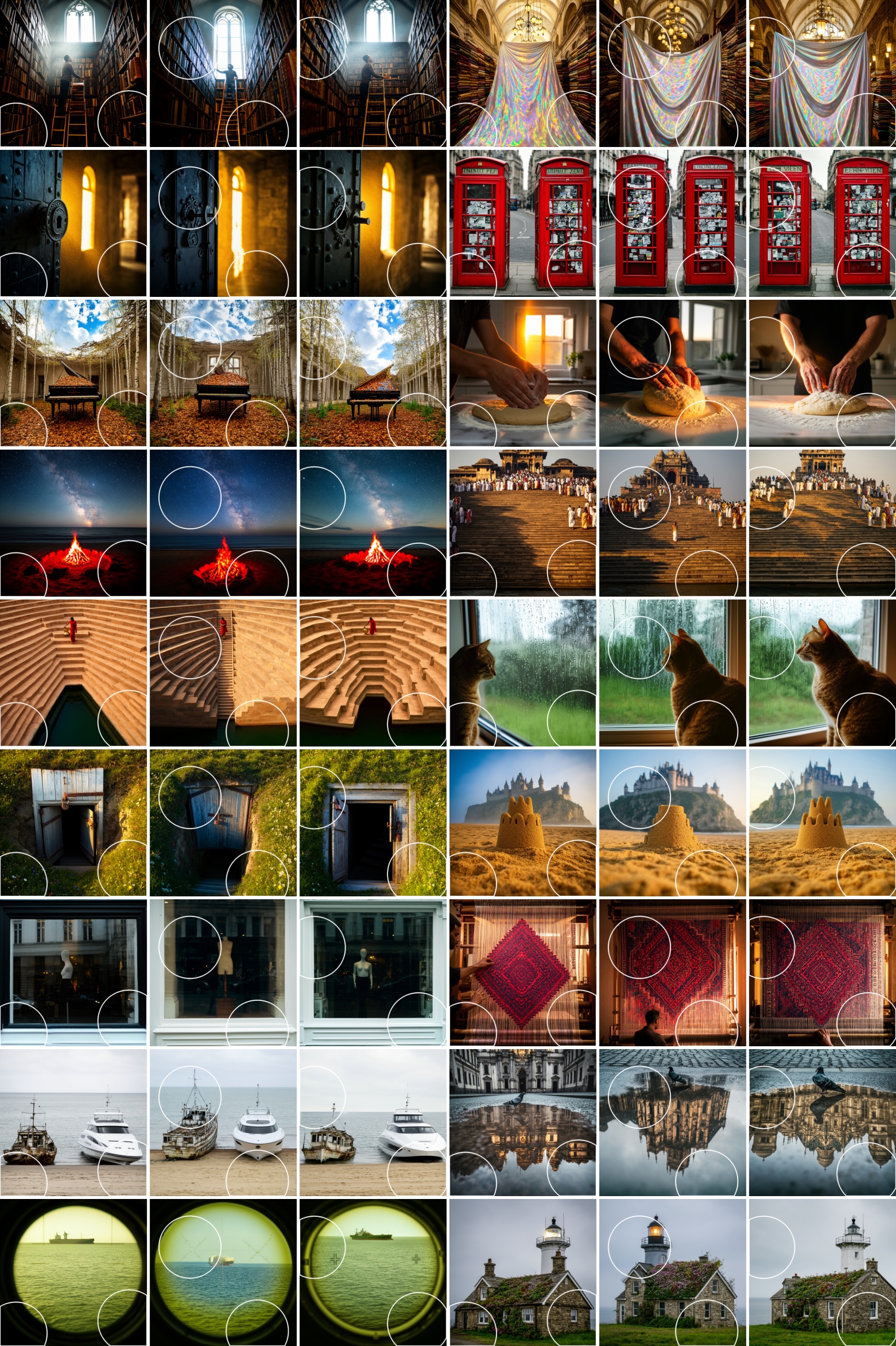}
    \caption{\textbf{Foveation mask containing multiple disjoint regions.} Foveated Diffusion generates coherent content with foveation masks containing multiple disjoint regions. Given the same prompt and noise seed, the generated images remain consistent and adhere to the prompt.}
    \label{fig:fov_pattern_multi_circle}
\end{figure}

We present additional results using the \textit{randomized-mask} model, where the high-resolution region varies in shape, size, and position, including foveation masks that contain multiple disjoint regions (Figs.~\ref{fig:fov_pattern_radius}--\ref{fig:fov_pattern_multi_circle}). Given the same prompt and noise seed, Foveated Diffusion generates coherent and consistent content independent of the foveation mask geometry; the mask only determines which regions are synthesized at high resolution.

Specifically, images generated with foveation masks containing multiple disjoint regions (Fig. \ref{fig:fov_pattern_multi_circle}) suggest that Foveated Diffusion could support multi-viewer scenarios with multiple gaze locations.

\clearpage

\clearpage

\clearpage

\clearpage
\subsection{Towards Saliency-guided Image Generation}
\label{supp_subsec:image_saliency}

\begin{figure}
    \centering
    \includegraphics[width=0.9\textwidth]{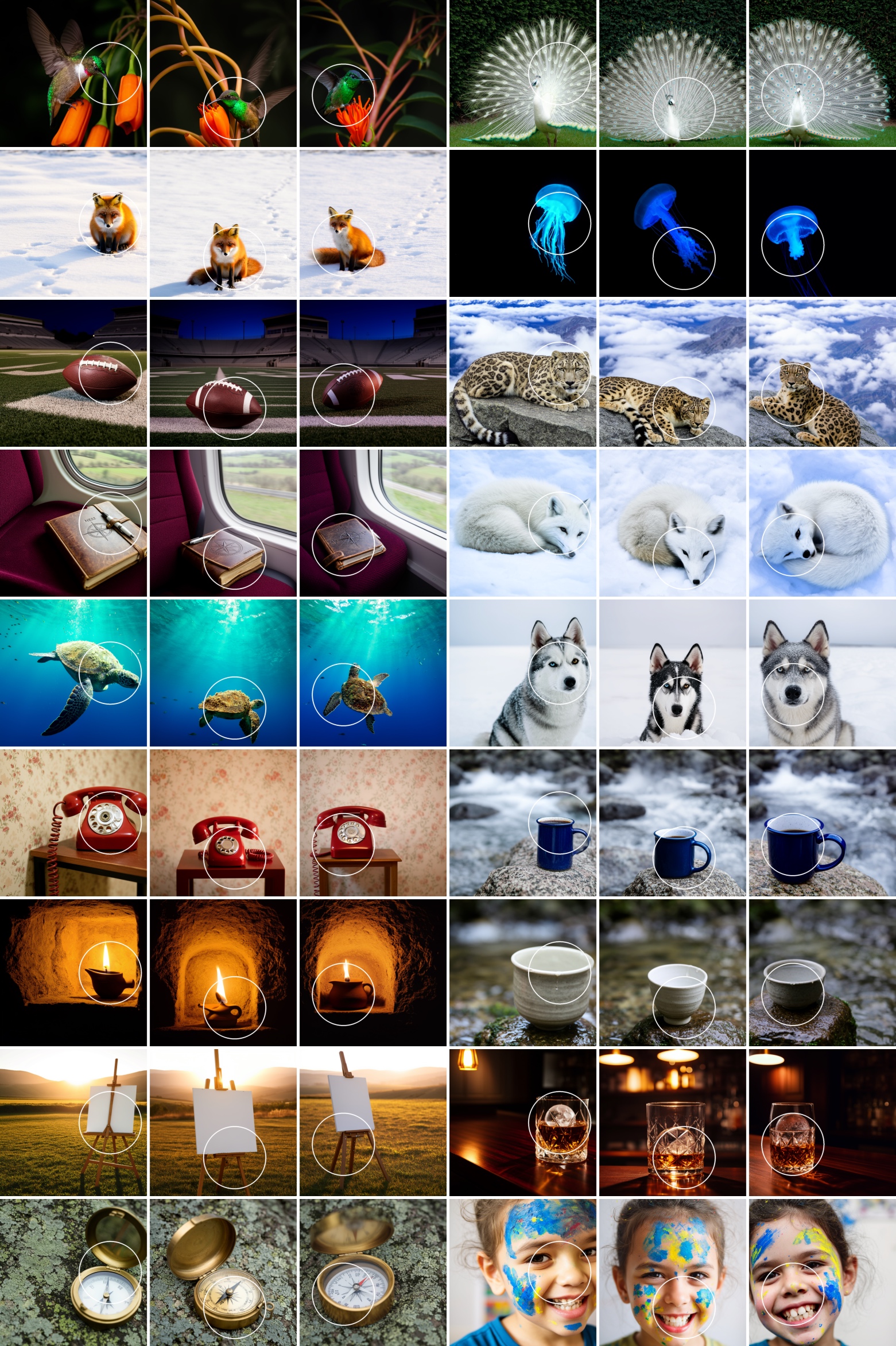}
    \caption{\textbf{Single-object saliency-guided generation.} Foveated Diffusion with saliency-guided training enables coarse, controllable single-object generation, where the salient object is approximately aligned with the center of the foveation mask.}
    \label{fig:saliency_single}
\end{figure}

\begin{figure}
    \centering
    \includegraphics[width=0.9\textwidth]{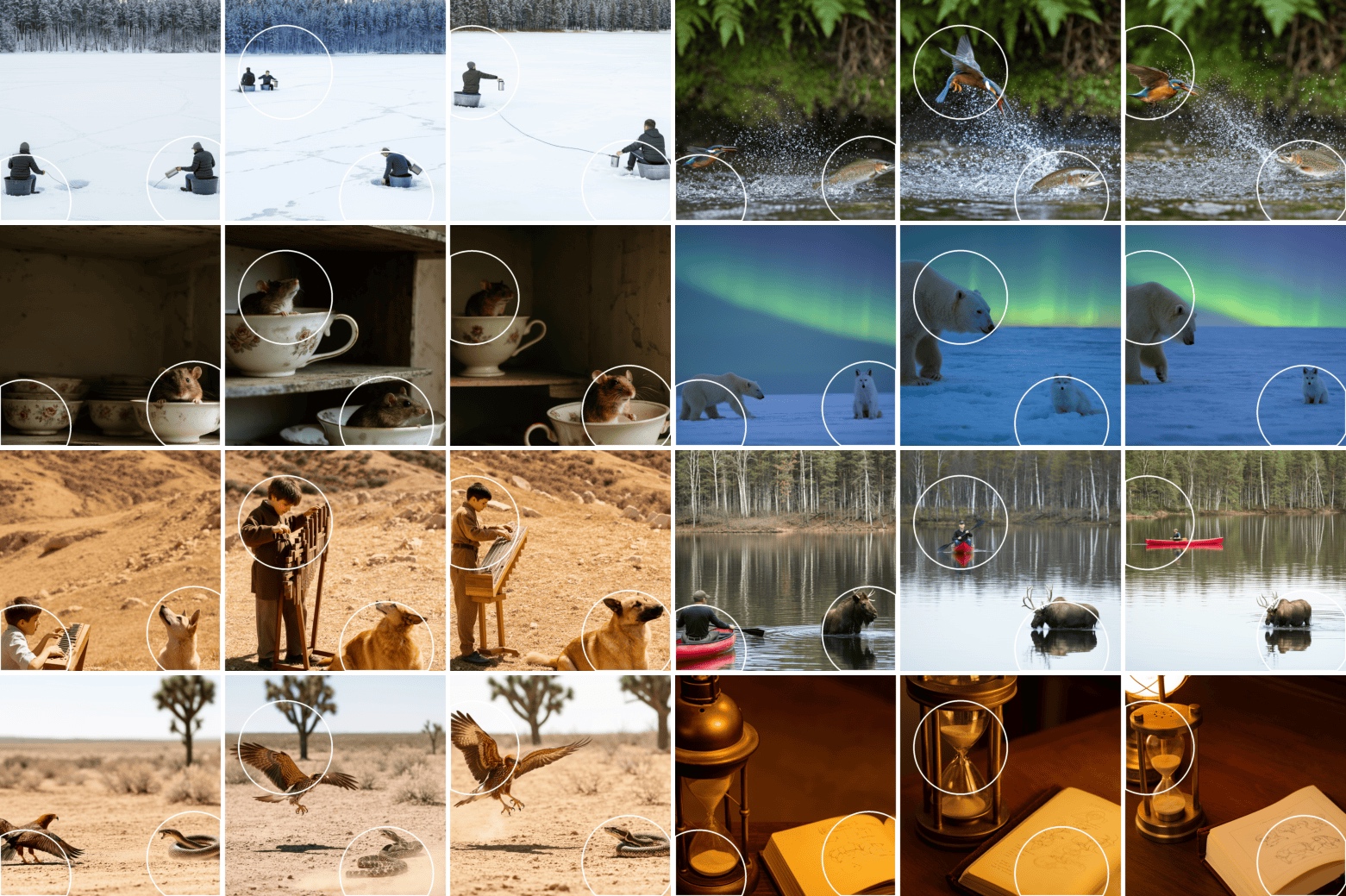}
    \caption{\textbf{Multi-object saliency-guided generation.} Foveated Diffusion with saliency-guided training also enables coarse, controllable multi-object generation, where salient objects approximately align with the centers of the disjoint regions in the foveation mask.}
    \label{fig:saliency_multi}
\end{figure}

\begin{figure}
    \centering
    \includegraphics[width=0.9\textwidth]{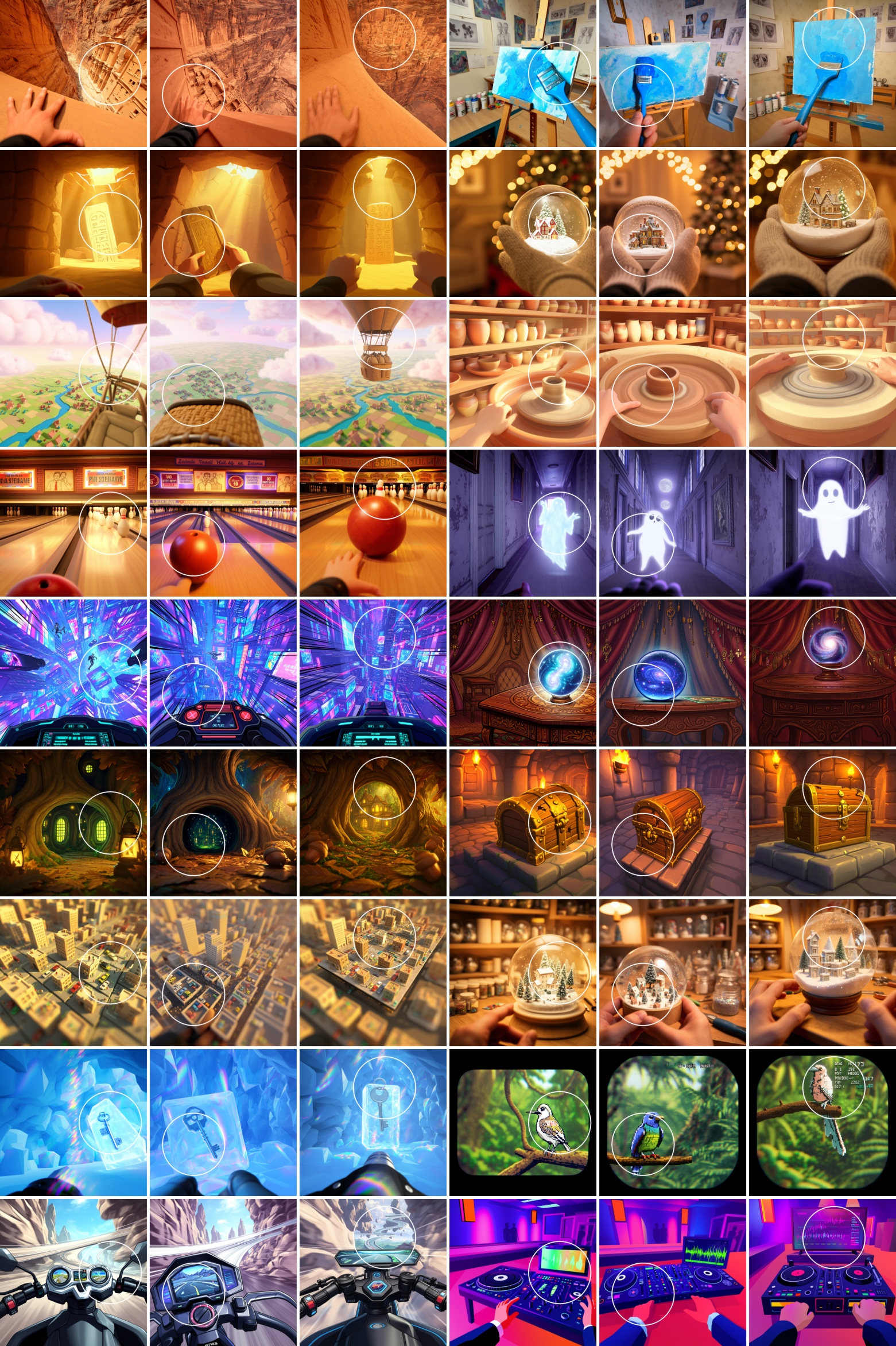}
    \caption{\textbf{Saliency-guided generation for immersive gaming.} Foveated Diffusion is well suited for immersive first-person generative gaming applications, where salient objects can be generated near the gaze-tracked location (fovea) and rendered in high resolution, while the remaining regions are rendered at lower resolution.}
    \label{fig:saliency_games}
\end{figure}

\begin{figure}
    \centering
    \includegraphics[width=0.9\textwidth]{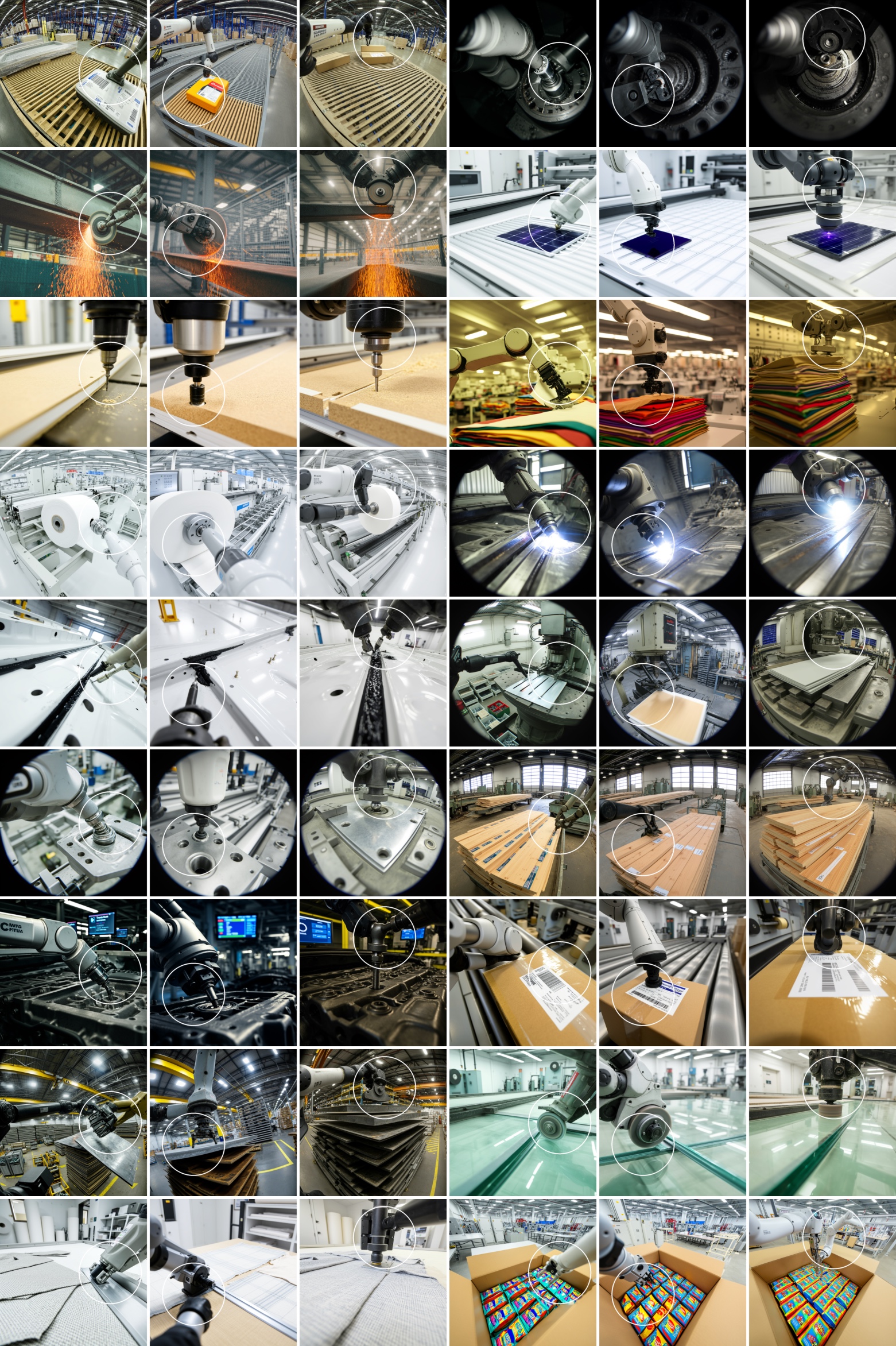}
    \caption{\textbf{Saliency-guided generation for robotics simulation.} 
    Foveated Diffusion is well suited for generative robotics simulation, where foveated imagery can be used for robotics policy learning. In this setting, only robot arms and manipulated objects are generated at high resolution, while the background is rendered at lower resolution to provide global context.}
    \label{fig:saliency_robotics}
\end{figure}

\begin{figure}
    \centering
    \includegraphics[width=0.9\textwidth]{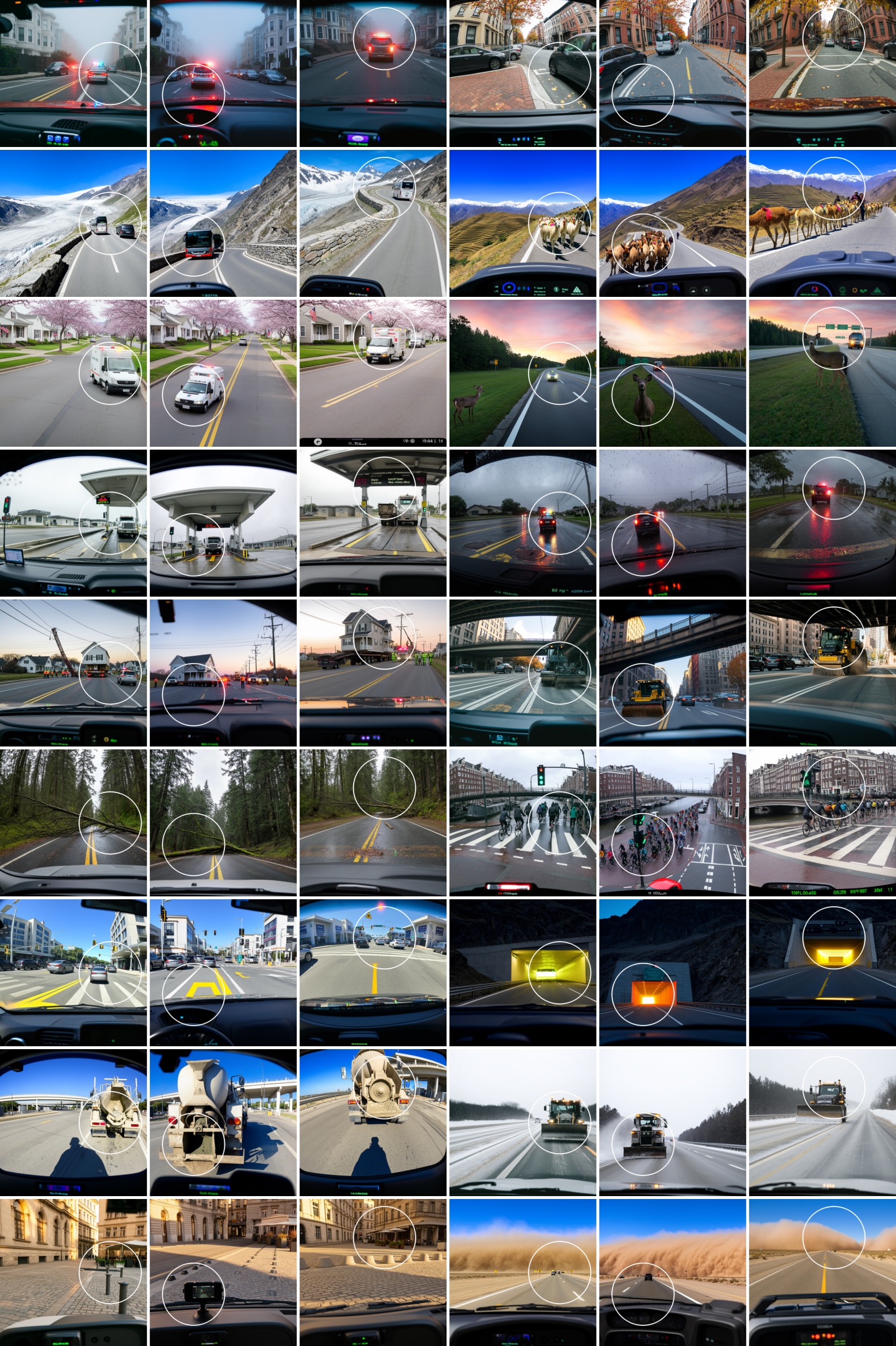}
    \caption{\textbf{Saliency-guided generation for autonomous vehicles.}  
    Foveated Diffusion is also well suited for generative autonomous driving simulation, where foveated imagery can be used for policy learning in self-driving systems. In this setting, only important objects in the scene (e.g., pedestrians, other vehicles, roadblocks) are generated at high resolution, while the background is rendered at lower resolution to provide global context.}
    \label{fig:saliency_av}
\end{figure}

We show additional Foveated Diffusion additional results using the \textit{saliency-guided} model in Figures \ref{fig:saliency_single} to \ref{fig:saliency_av}. Compared to the randomized-mask model, the saliency-guided model generates images in which salient objects are aligned with the high-resolution regions defined by the foveation mask. Furthermore, the saliency-guided model also natively supports controllable multi-object generation when the foveation mask contains multiple disjoint high-resolution regions.

In Figures \ref{fig:saliency_games}--\ref{fig:saliency_av}, we illustrate potential applications of saliency-guided Foveated Diffusion, including immersive VR gaming (Fig.~\ref{fig:saliency_games}), generative robotics (Fig.~\ref{fig:saliency_robotics}), and autonomous driving simulation for robotics policy learning (Fig.~\ref{fig:saliency_av}). Foveated Diffusion is particularly well suited for these scenarios because only the most salient objects need to be rendered in high resolution (e.g., wielded objects in VR games, robot arms and manipulated objects, and pedestrians or vehicles in dashcam scenes), while the remaining regions can be rendered at lower resolution.

\clearpage

\clearpage

\clearpage

\clearpage

\clearpage

\clearpage
\subsection{Towards Bounding-box-guided Image Generation} 
\label{supp_subsec:image_bbox}

Similar to saliency-guided visual generation, we adapt Foveated Diffusion for \textit{bounding-box-guided} visual generation. We use the Ultralytics software library \cite{jocher2023ultralyticsyolo}, which integrates multiple YOLO models, for bounding box detection.

\begin{figure}
    \centering
    \includegraphics[width=0.9\textwidth]{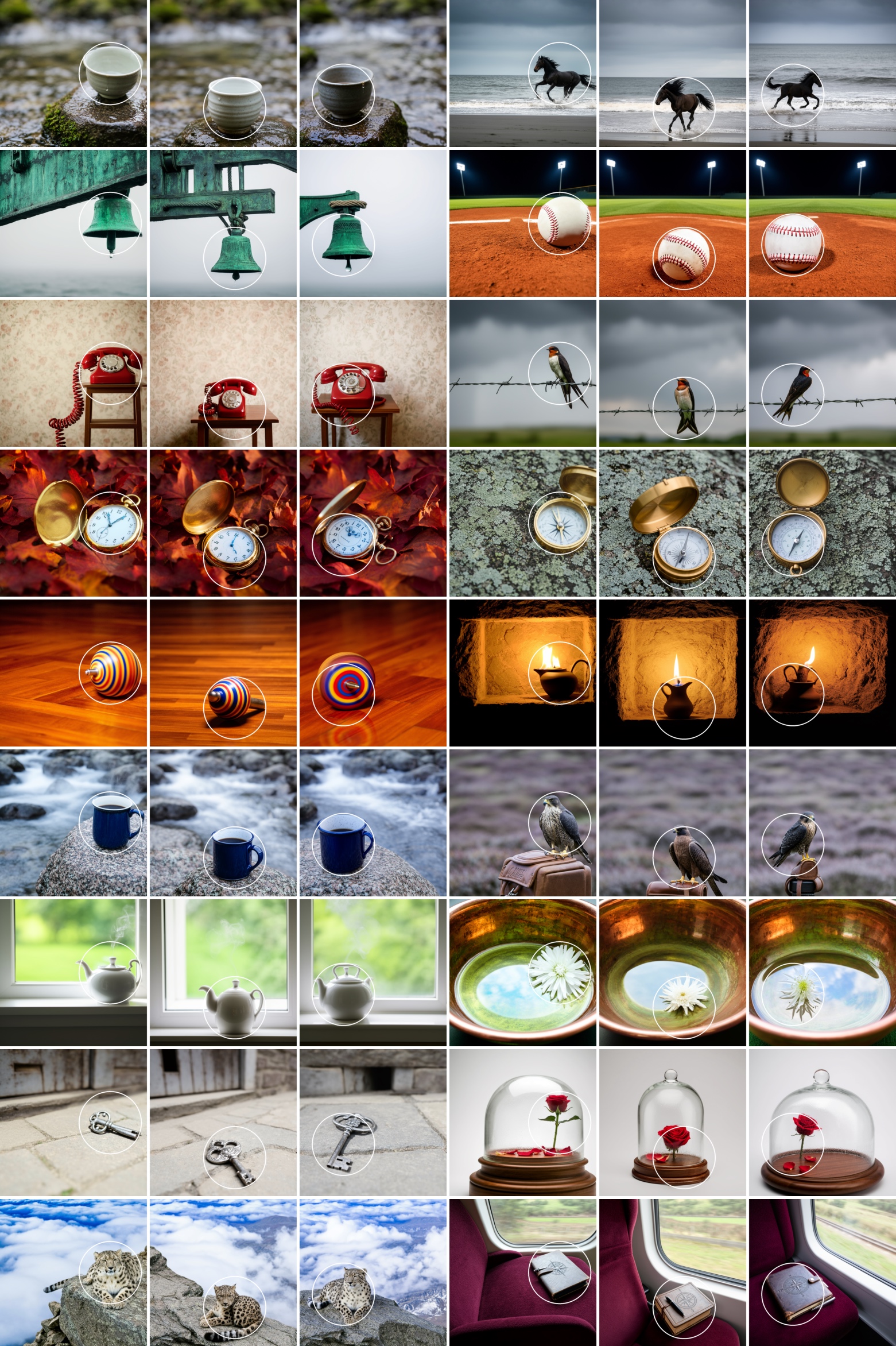}
    \caption{\textbf{Single-object bounding-box-guided generation.} Foveated Diffusion with bounding-box-guided training enables coarse, controllable single-object generation, where the salient object is approximately constrained within the foveation mask.}
    \label{fig:bbox_single}
\end{figure}

\begin{figure}
    \centering
    \includegraphics[width=0.9\textwidth]{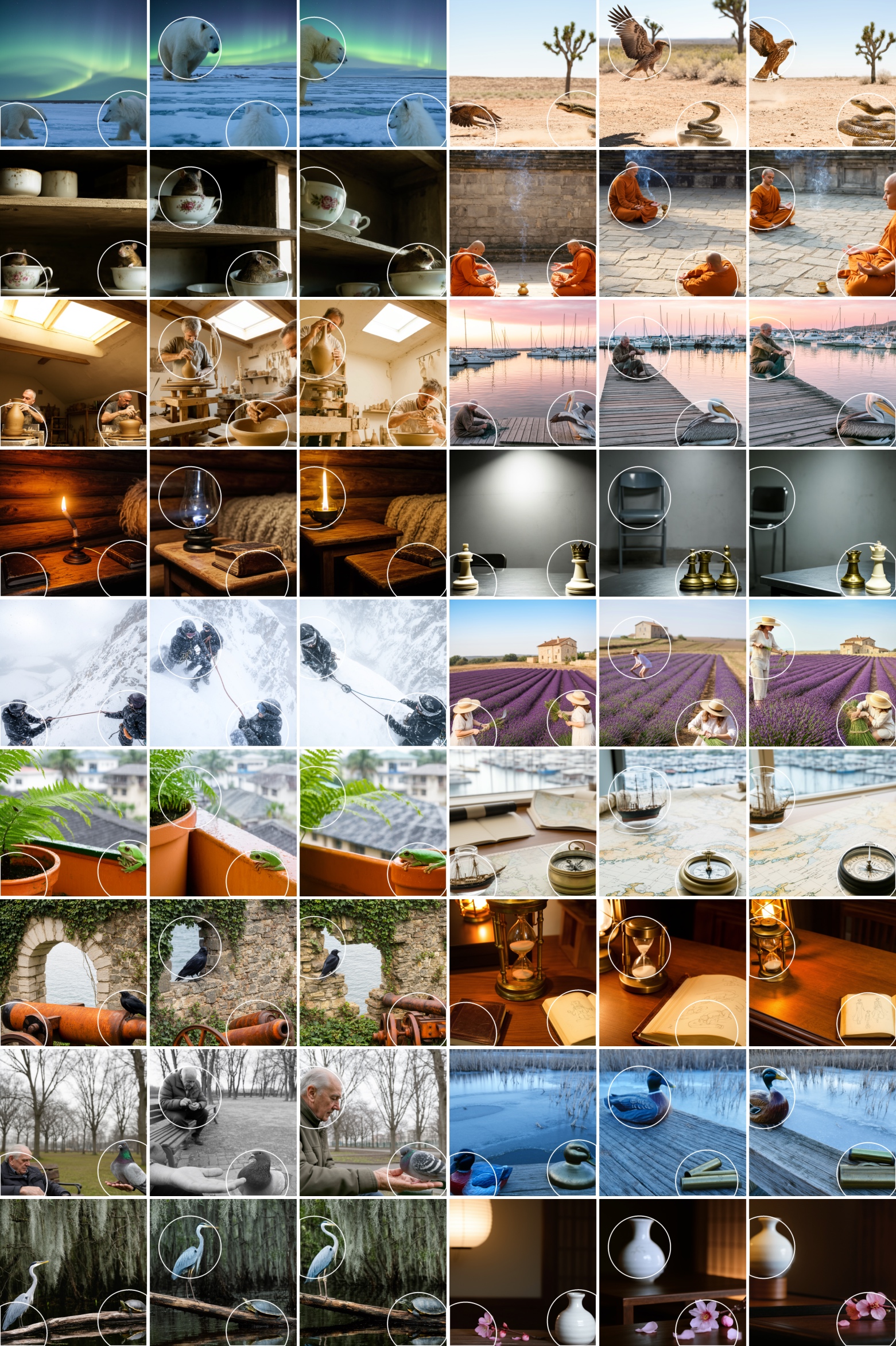}
    \caption{\textbf{Multi-object bounding-box-guided generation.} Foveated Diffusion with bounding-box-guided training enables coarse, controllable multi-object generation, where salient objects are approximately constrained within the disjoint regions of the foveation mask.}
    \label{fig:bbox_multi}
\end{figure}

\begin{figure}
    \centering
    \includegraphics[width=0.9\textwidth]{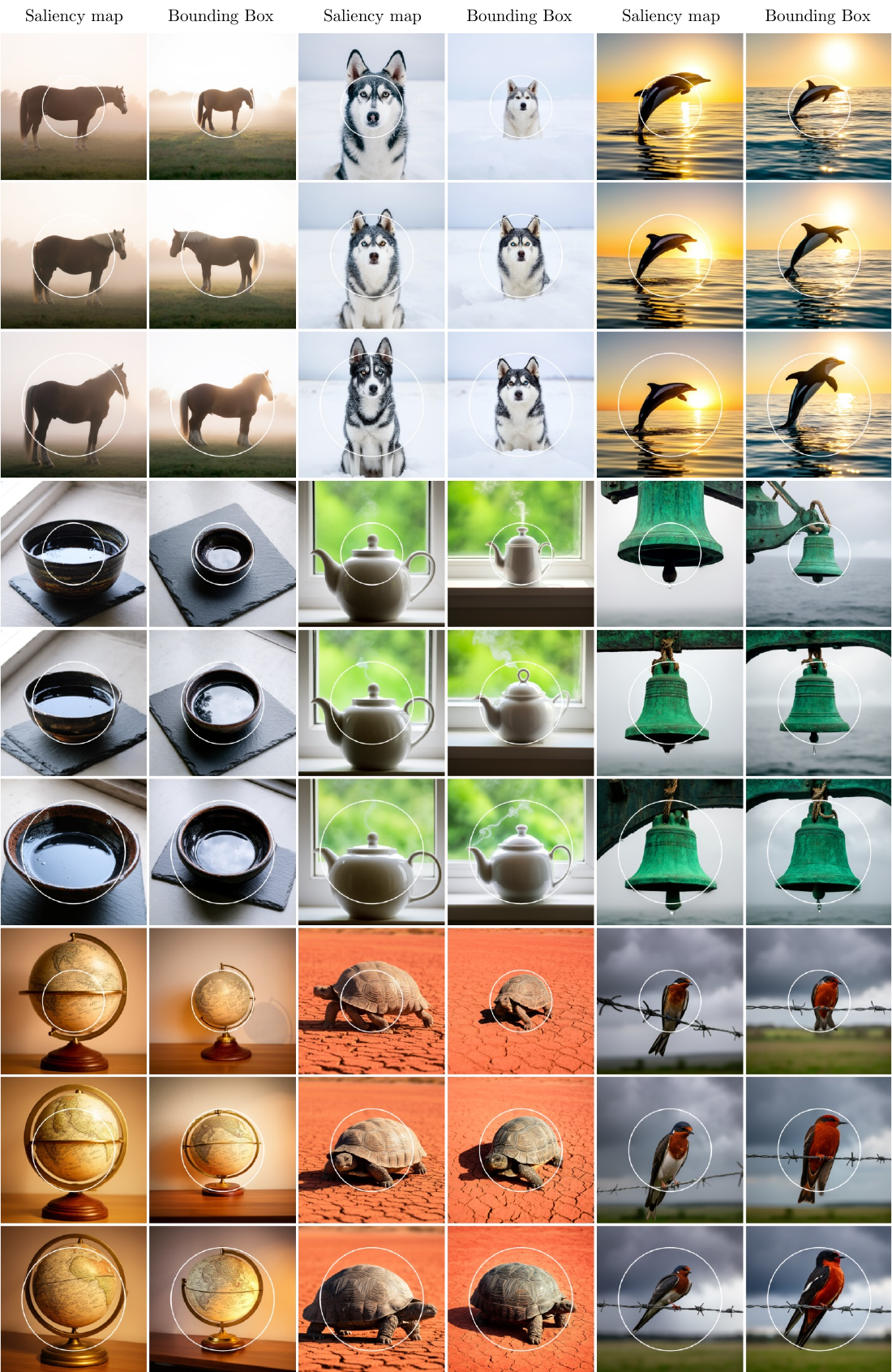}
    \caption{\textbf{Bounding-box-guided generation and saliency-guided generation.} We compare bounding-box-guided and saliency-guided generation. Because bounding-box-derived masks precisely delineate object contours, the bounding-box-guided model generates entire objects within the high-resolution region. In contrast, the saliency-guided model aligns only the most salient parts of objects with the center of the high-resolution region. }
    \label{fig:bbox_vs_saliency}
\end{figure}

As shown in Figures \ref{fig:bbox_single} to \ref{fig:bbox_vs_saliency}, the bounding-box-guided model successfully generates objects within the foveation boundary. Similar to the saliency-guided model, the bounding-box-guided model inherently enables controllable multi-object generation when the foveation mask comprises multiple disjoint high-resolution regions.

The difference between the saliency-guided and bounding-box-guided models is subtle but informative. Bounding boxes explicitly delineate object contours, encouraging the model to generate entire objects within, or closely aligned to, the foveal region (Fig. \ref{fig:bbox_vs_saliency}). In contrast, the saliency-guided model aligns only the most salient portions of objects with the fovea, rather than enforcing full-object containment. Notably, this behavior is not imposed by any architectural modification or specialized algorithm. This behavior arises purely from data construction, namely how foveation masks structure the interaction between high- and low-resolution tokens during training, highlighting the generality of our Foveated Diffusion framework.

\clearpage

\clearpage

\clearpage

%% file: supp_sections/video_qualitative.tex
We include selected video generation results on our project website \footnote{\url{https://bchao1.github.io/foveated-diffusion/}}.

\subsection{Extended Video Generation Baseline Comparisons}
\label{supp_subsec:video_baseline}

We provide extended baseline comparisons against full high-resolution video generation and na\"ive mixed-resolution video generation using our \textit{randomized mask model}. While the foveation mask for image generation is defined as a circle with a randomized center and radius, video generation requires temporal coherence. To achieve this, we sample three key control points with randomized spatial coordinates and radii across the video sequence. We then apply cubic spline interpolation to these points to generate a smooth, continuous foveation trajectory for the duration of the video, ensuring the high-resolution window moves fluidly across frames. We show both 480p and 720p generation results to show the generality of our model.

Our method consistently outperforms the na\"ive mixed-resolution baseline, generating coherent and consistent content with consistent structure and scale without color distortions, whereas the mixed-resolution baseline exhibits significant artifacts.

\subsection{Video Generation with Different Foveation Patterns} 
\label{supp_subsec:video_fov}

We present additional video generation results using the \textit{randomized-mask} model, where the high-resolution region follows different randomized spline trajectories that vary in position and size across frames. Given the same prompt and noise seed, Foveated Diffusion generates coherent and consistent content independent of the foveation mask trajectory; the mask only determines which regions are synthesized at high resolution.

\subsection{Towards Saliency-guided Video Generation}
\label{supp_subsec:video_saliency}

We show additional Foveated Diffusion additional results using the \textit{saliency-guided} model. The saliency-guided model generates videos in which salient objects are aligned with the high-resolution regions defined by the foveation mask trajectory.

We illustrate potential applications of saliency-guided Foveated Diffusion, including immersive VR gaming and generative robotics and autonomous driving simulation for robotics policy learning. Foveated Diffusion is particularly well suited for these scenarios because only the most salient objects need to be rendered in high resolution (e.g., wielded objects in VR games, robot arms and manipulated objects, and pedestrians or vehicles in dashcam scenes), while the remaining regions can be rendered at lower resolution.

%% file: supp_sections/discussion.tex
\begin{figure*}
    \centering
    \includegraphics[width=\textwidth]{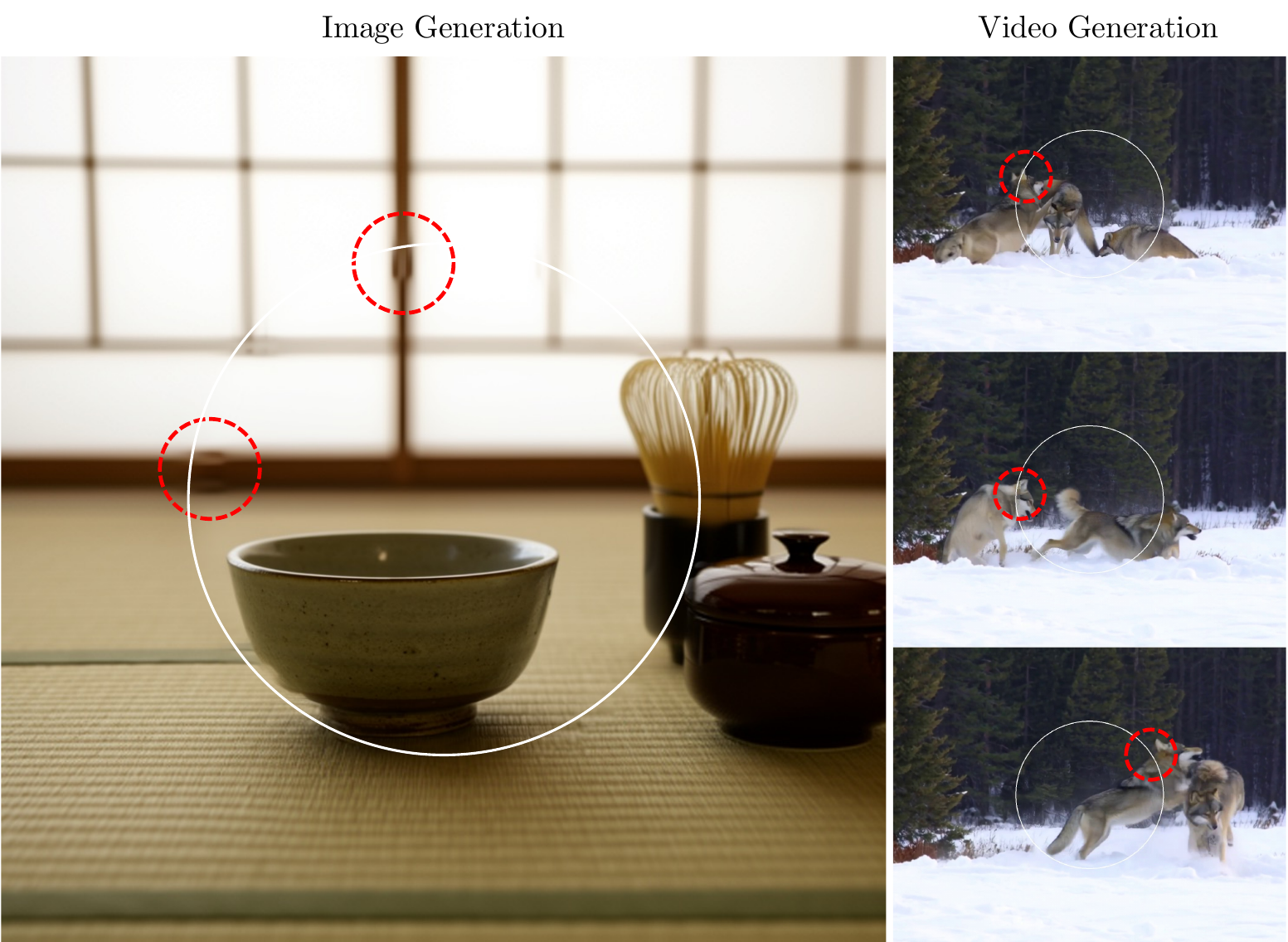}
    \caption{\textbf{Foveated Diffusion artifacts.} We delineate the foveation border with a white circular outline. The red dashed lines indicate regions with blending artifacts.}
    \label{fig:artifacts}
\end{figure*}

Foveated Diffusion occasionally exhibits color or discontinuity artifacts near foveation boundaries, as shown in Fig.~\ref{fig:artifacts}. We attribute these artifacts to the final VAE decoding and alpha-blending step between low- and high-resolution regions. We believe these artifacts could be mitigated by adapting the VAE to directly decode mixed-resolution tokens, thereby avoiding separate decoding and blending of low- and high-resolution regions.